\documentclass[10pt,twocolumn,letterpaper]{article}

\usepackage[pagenumbers]{cvpr} 
\usepackage[accsupp]{axessibility}

\usepackage{times}
\usepackage{epsfig}
\usepackage{graphicx}
\usepackage{amsmath}
\usepackage{color}
\usepackage{amssymb}
\usepackage{lipsum}
\usepackage{algorithm}
\usepackage{algpseudocode}

\usepackage{url}
\usepackage{xcolor, colortbl}
\usepackage{mathtools}
\usepackage{tabularx}
\usepackage{multirow}
\usepackage{enumitem}
\usepackage{bbm}
\usepackage{wrapfig}
\usepackage{setspace}
\RequirePackage{fix-cm}

\def\ie{\emph{i.e.}}
\def\eg{\emph{e.g.}}
\def\etal{\emph{et al.}}

\newcolumntype{C}[1]{>{\centering\let\newline\\\arraybackslash\hspace{0pt}}p{#1}}

\newcommand{\Fig}[1]{Fig.~\ref{fig:#1}}

\newcommand{\sftype}[1]{{\textsf{\small #1}}}
\newcommand{\sftypes}[1]{{\textsf{\tiny #1}}}
\newcommand{\expnum}[2]{{#1}\mathrm{e}{-#2}}

\usepackage{booktabs}
\usepackage[pagebackref,breaklinks,colorlinks]{hyperref}

\usepackage[capitalize]{cleveref}
\crefname{section}{Sec.}{Secs.}
\Crefname{section}{Section}{Sections}
\Crefname{table}{Table}{Tables}
\crefname{table}{Tab.}{Tabs.}

\begin{document}

\title{FIFO: Learning Fog-invariant Features for Foggy Scene Segmentation}

\author{Sohyun Lee\\
GSAI, POSTECH\\
{\tt\small lshig96@postech.ac.kr}
\and
Taeyoung Son\thanks{This work was done while Taeyoung Son was in POSTECH.}\\ 
NALBI\\
{\tt\small taeyoung@nalbi.ai}
\and
Suha Kwak\\
GSAI, POSTECH\\
{\tt\small suha.kwak@postech.ac.kr} \\
\vspace{-7mm}
\and
{\small \url{http://cvlab.postech.ac.kr/research/FIFO/}}
}
\maketitle

\begin{abstract}

        Robust visual recognition under adverse weather conditions is of great importance in real-world applications. 
        In this context, we propose a new method for learning semantic segmentation models robust against fog.
        Its key idea is to consider the fog condition of an image as its style and close the gap between images with different fog conditions in neural style spaces of a segmentation model.
        In particular, since the neural style of an image is in general affected by other factors as well as fog, we introduce a \emph{fog-pass filter} module that learns to extract a fog-relevant factor from the style.
        Optimizing the fog-pass filter and the segmentation model alternately gradually closes the style gap between different fog conditions and allows to learn fog-invariant features in consequence.
        Our method substantially outperforms previous work on three real foggy image datasets.
        Moreover, it improves performance on both foggy and clear weather images, while existing methods often degrade performance on clear scenes.

\end{abstract}
\section{Introduction}\label{sec:intro}

We have witnessed great advances in semantic segmentation for the last decade.
However, most of existing models and datasets focus merely on improving accuracy under controlled environments, without considering image degradation caused by adverse weather conditions (\eg, fog, rain, and snow), over- and under-exposure, motion blur, sensor noise, \emph{etc}.
The robustness of semantic segmentation models against these factors is of great importance in safety-critical applications
and recently has gained increasing attention~\cite{Sakaridis_2018_ECCV,dai2020curriculum,Sakaridis_2019_ICCV,bijelic2020seeing,Zendel_2018_ECCV,Sakaridis_2018_IJCV,son2020urie,choi2021robustnet}.

\begin{figure}
    \centering
    \includegraphics[width=0.47\textwidth]{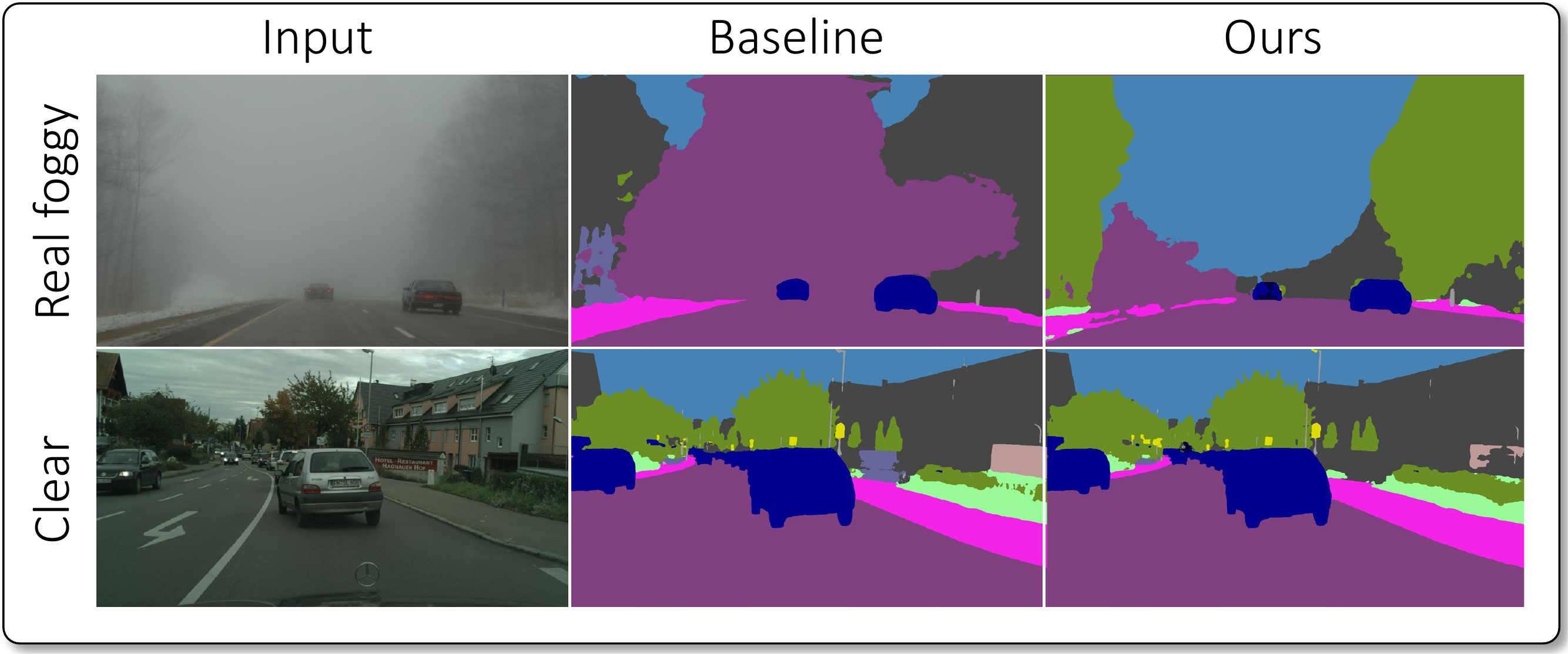}
    \vspace{-2mm}
    \caption{
    A summary of our results. 
    Predictions of FIFO are accurate for both clear weather and real foggy images while the baseline, an ordinary segmentation model trained on clear weather images, fails to handle foggy images.
    }
    \label{fig:teaser}
    \vspace{-4mm}
\end{figure}

Motivated by this, we study semantic segmentation of \emph{foggy} scenes, whose goal and results are illustrated in Fig.~\ref{fig:teaser}.
The task is challenging since fog often damages visibility of images seriously, leading to substantial performance degradation.
Attaching a fog removal network to the front of an existing model is not always useful for mitigating this issue~\cite{Pei_2018_ECCV,Sakaridis_2018_IJCV} as well as being heavy in computation and memory.
The other reason for the difficulty is the absence of fully annotated data for the task.
Collecting a large set of foggy scenes is not straightforward since they can be captured under only a specific condition, and it is hard to label them due to their limited visibility. 

\begin{figure*}[t]
    \centering
    \vspace{-2mm}
    \includegraphics[width=0.95\linewidth]{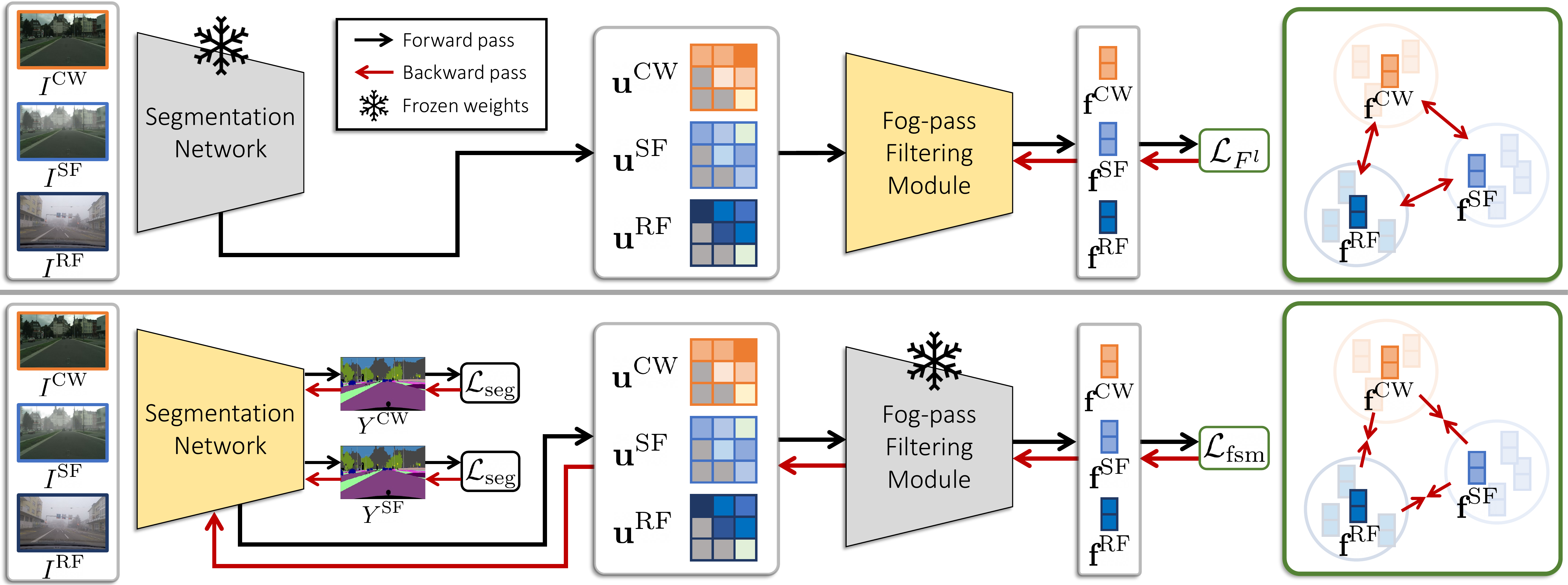}
    \vspace{-3mm}
\caption{
Overall pipeline of FIFO.
For each iteration of training, the fog-pass filtering module and the segmentation network are updated alternately. 
(\emph{top}) Given Gram matrices of feature maps of the segmentation network as input, the fog-pass filtering module learns to extract fog factors
so that fog conditions of images are discriminated by their fog factors. 
(\emph{bottom}) The segmentation network is trained by reducing the gap between fog factors of images with different fog conditions as well as by the segmentation loss.
Note that the fog-pass filters are auxiliary modules used only in training.
} \label{fig:pipeline}
\vspace{-2mm}
\end{figure*}

Existing methods~\cite{Sakaridis_2018_IJCV,Sakaridis_2018_ECCV,dai2020curriculum} tackle these issues through synthetic foggy image datasets, which are obtained by applying realistic fog effects to fully annotated clear weather images and are used for supervised learning of semantic segmentation.
Furthermore, they introduce curriculum learning approaches~\cite{Sakaridis_2018_ECCV,dai2020curriculum} that gradually adapt a model from light synthetic fog to dense real fog using unlabeled real foggy images additionally.
Although these methods have achieved impressive robustness, there remains room for further improvement in that their training strategies are limited to ordinary supervised learning.
In addition, the curriculum adaptation demands external modules to control the fog density of real foggy images in training, and tends to make the final model biased to foggy scenes;
it thus requires extra computation and additional hyper-parameters in training, and
often degrades performance on clear images.

To resolve the above issues, we propose a new method that learns Fog-Invariant features for FOggy scene segmentation, dubbed FIFO.
Its overall pipeline is illustrated in Fig.~\ref{fig:pipeline}.
FIFO considers the fog condition of an image as its style, ideally independent of its content, and aims to learn a segmentation model insensitive to fog style variation of input image. 
To this end, we first define three different domains of training images, \ie, clear weather (\sftype{CW}), synthetic fog (\sftype{SF}), and real fog (\sftype{RF}), where images of the first two domains are labeled while those of the last one are not.
FIFO then encourages the segmentation network to close the style discrepancy between different fog domains in feature spaces so that it learns fog-invariant features.

Then the success of FIFO depends heavily on the quality of the fog style representation.
Unfortunately, existing style representation schemes~\cite{gatys2016image,Instancenorm} are not desirable for our task since they are manually designed to capture the holistic style of an image that is affected also by factors other than fog (\eg, when and where the image was taken) and even the content of the image~\cite{choi2021robustnet};
the direct use of these neural styles thus introduce side-effects like content alteration and result in suboptimal solutions consequently.

To address this issue, we present \emph{fog-pass filters}, learnable modules that take an ordinary neural style---the Gram matrix of a feature map~\cite{gatys2016image} as input and extract only a fog-relevant information from the style precisely in the form of embedding vectors, called \emph{fog factors}.
In particular, they learn to draw fog factors of the same domain together and hold those of different domains apart so that they discriminate fog conditions of input images through their fog factors.
The segmentation model is in turn encouraged to reduce the gap between fog factors of images from different domains during training. 
The alternating optimization of the fog-pass filter and the segmentation network gradually closes the fog style gap between different domains and eventually leads to fog-invariant features of the segmentation network.
Note that the fog-pass filters are auxiliary modules for training only, thus not required in testing.

FIFO has advantages over the previous work~\cite{Sakaridis_2018_IJCV,Sakaridis_2018_ECCV,dai2020curriculum} in terms of both simplicity in training and efficacy in testing.
Unlike the previous work, FIFO does not need to control fog density levels of synthetic and real foggy images in training, thus allowing end-to-end learning of a segmentation model with fewer hyper-parameters.
More importantly, for the same reason, it demands no extra module for estimating and manipulating fog density of real foggy images in training.
Regarding the effectiveness, FIFO clearly outperforms all existing records and improves performance on both foggy and clear weather domains, while existing methods often degrade performance on clear scenes.

\section{Related Work} \label{sec:relatedwork}

\noindent\textbf{Semantic Foggy Scene Segmentation.}
Previous work~\cite{Sakaridis_2018_IJCV,Sakaridis_2018_ECCV,dai2020curriculum} has developed fog simulators that are applied to clear images with full annotations to obtain labeled synthetic foggy images.
Since supervised learning on the synthetic data limits performance due to the visual gap between synthetic and real foggy images,
recent methods~\cite{dai2020curriculum, Sakaridis_2018_ECCV} further employ curriculum learning to gradually adapt a model from light synthetic fog to dense real fog.
However, the curriculum adaptation often degrades performance on clear weather images and demands extra modules to control density levels of real foggy images during training.

\noindent\textbf{Image Dehazing.}
Fog damages visibility of image, and accordingly, degrades visual recognition performance substantially.
Numerous dehazing algorithms have been proposed so far to restore latent clean image from foggy input~\cite{fattal2008single,he2010single,fattal2014dehazing,berman2016non,Li_2017_ICCV,zhang2018densely,Chen_2019_CVPR,Liu_2019_ICCV}.
However, they are usually too heavy in computation to be attached to the front of recognition models.
Further, recent studies~\cite{Pei_2018_ECCV,Sakaridis_2018_IJCV} suggest that most dehazing models do not help improve recognition performance on foggy scenes.
Hence, instead of having a separate dehazing module, our work learns a segmentation model whose features are invariant to fog condition.

\noindent\textbf{Robustness.}
Robustness of recognition models against adverse conditions has been actively investigated due to its importance in real-world applications~\cite{goodfellow2014explaining,Hendrycks2019_ImageNetC,Zendel_2018_ECCV,hendrycks2018deep,Hu_2019_CVPR,acdc,franchi,porav2020rainy,jin2021raidar}, and a variety of methods have been proposed to improve robustness~\cite{son2020urie,schneider2020improving,shi2020informative,Wang_VRR2016,vpgnet,choi2021robustnet}.
FIFO shares a similar idea with RobustNet~\cite{choi2021robustnet}, which regards adverse condition of input as its style and removes the effect of photometric transform of input from a neural style representation of a model so that the model becomes invariant to the transform.
Compared to RobustNet, FIFO more explicitly quantifies the effect of adverse condition through a learnable module (\ie, fog-pass filter), so is able to more precisely manipulate the adverse effects during training. 

\noindent\textbf{Style Transfer.}
Neural style transfer has been studied to comprehend the style of an image apart from its content~\cite{berger2016incorporating,ulyanov2016texture,gatys2017controlling,huang2017arbitrary,dumoulin2016learned,li2017universal,song2019etnet}.
In particular, a seminal work~\cite{gatys2016image} studied the Gram matrix of a feature map as a neural style representation and showed that the style of an image can be transferred to another by approximating its Gram matrix;
the efficacy of Gram matrix has been proven further in later studies~\cite{johnson2016perceptual,luan2017deep}.
Also, Li~\etal~\cite{li2017demystifying} proved that matching Gram matrices is useful for domain adaptation since it is equivalent to minimizing maximum mean discrepancy between domains.
However, we found that Gram matrix is not appropriate as-is for quantifying the effect of fog in our task since it is affected other style factors or even content~\cite{choi2021robustnet} as well as fog.
We thus introduce the fog-pass filter to precisely capture only a fog-relevant factor from the Gram matrix of a feature map.

\noindent\textbf{Unsupervised Domain Adaptation (UDA).}
Our work is also relevant to UDA since both adapt models to an unlabeled target domain. 
UDA methods for semantic segmentation can be categorized by the level at which adaptation is performed: Input-level~\cite{murez2018image, hoffman2018cycada, pizzati2020domain, kang2020pixel}, feature-level~\cite{wang2020classes, AdaptSegNet, luo2019significance}, and output-level~\cite{zou2018unsupervised, zou2019confidence, luo2019taking}.
FIFO is related in particular to the feature-level adaptation that learns domain-invariant features.
Most of existing methods in this category~\cite{wang2020classes, AdaptSegNet, luo2019significance} train a discriminator together with a segmentation model so that the discriminator maximizes a discrepancy between source and target domains while the segmentation model learns to minimize the discrepancy.
FIFO shares a similar idea with these methods, but as will be demonstrated, closing the gap between fog factors in FIFO is 
more effective than fooling a fog domain classifier in fog-invariant feature learning.
\section{Configuration of Training Data} \label{sec:dataset}

Training images for FIFO are categorized into three different domains according to their fog types: clear weather (\sftype{CW}), synthetic fog (\sftype{SF}), and real fog (\sftype{RF}). 
For \sftype{CW} images, we adopt the Cityscapes dataset~\cite{cityscapes}, which is fully annotated for supervised learning of semantic segmentation. 
Meanwhile, as \sftype{SF} images, we utilize the Foggy Cityscapes-DBF dataset~\cite{Sakaridis_2018_ECCV}, which is constructed by simulating realistic fog effects on images of the Cityscapes dataset, thus also fully annotated.
Finally, \sftype{RF} images are taken from the Foggy Zurich dataset~\cite{Sakaridis_2018_ECCV}, which is a collection of unlabeled foggy scenes captured in the real world.

Note that the way FIFO uses the two foggy image datasets is different from that of existing methods~\cite{Sakaridis_2018_ECCV,dai2020curriculum}.
First, in the Foggy Cityscapes-DBF dataset, FIFO fixes the density level of synthetic fog by a single value (\ie, the attenuation coefficient $\beta=0.005$) and utilizes the entire dataset.
On the other hand, the previous work adopts only a refined, high-quality subset of the dataset and varies the fog level during training for the curriculum learning.
Second, FIFO utilizes the Foggy Zurich dataset as a whole, whereas the previous work divides it into multiple sets of different density levels using extra modules that estimate the fog density of the images. 
These differences allow the pipeline of FIFO to be more straightforward and concise. 

\section{Proposed Method} \label{sec:method}

\begin{figure}[t]
    \centering
    \vspace{-2mm}
    \includegraphics[width=1\linewidth]{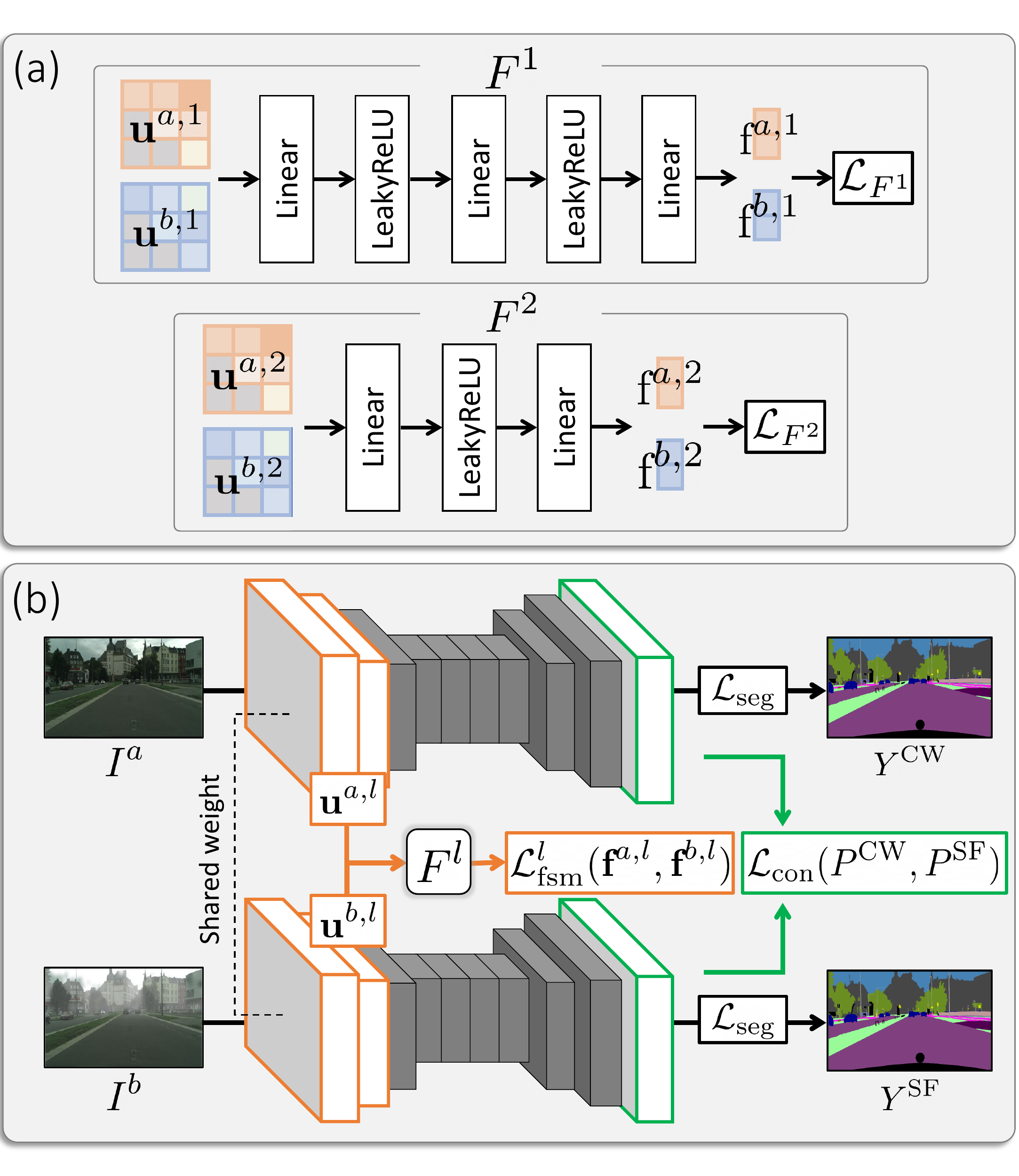}
    \vspace{-8mm}
\caption{
    A schematic of FIFO. 
    (a) The fog-pass filtering modules.
    Each of them takes as input the upper triangular part of a Gram matrix and returns a fog factor. 
    The loss pulls or pushes a pair of fog factors according to the equivalence of their fog conditions. 
    (b) Training of the segmentation network with the frozen fog-pass filters. 
    Given a pair of images with different fog conditions as input, the network is trained by closing the gap between their fog factors and that between their segmentation predictions as well as by the ordinary segmentation loss.
} \label{fig:detail}
\vspace{-3mm}
\end{figure}

The training procedure for the fog-pass filters and the segmentation network is illustrated in Fig.~\ref{fig:pipeline} and their structures are presented in Fig.~\ref{fig:detail}.
The segmentation network is first pretrained on the Cityscapes dataset~\cite{cityscapes}, and fog-pass filters initialized randomly are attached to different feature maps of the network.
Thereafter, on the dataset of the three fog domains introduced in Sec.~\ref{sec:dataset}, the two parts of FIFO are trained alternately per mini-batch, except for the first 5K iterations where the fog-pass filters are solely trained to avoid cold-start. 
Note that the fog-pass filters are used only in training for fog-invariant feature learning of the segmentation network.
In other words, FIFO imposes no additional inference-time complexity since it employs only the segmentation network for testing.

To construct a mini-batch, we randomly sample the same number of images from \sftype{CW} and \sftype{RF} domains, and choose \sftype{SF} counterparts of the sampled \sftype{CW} images.
Given such a mini-batch, the fog-pass filters are learned to draw fog factors of the same fog domain together and hold those of different domains apart so that they discriminate input according to its fog domain.
On the other hand, the segmentation network is optimized for closing the distance between fog factors of different domains as well as for minimizing the ordinary segmentation loss.
This alternating optimization closes the fog style gap between different domains precisely, leading to fog-invariant features.

The remaining part of this section first gives details of training the fog-pass filtering modules and the segmentation network, then empirically verifies the key ideas of FIFO.

\subsection{Fog-pass Filtering Modules} \label{sec:fog-pass_filter}

Instead of a raw feature map of the segmentation network, a fog-pass filtering module takes a holistic style representation of the feature map as input in order to focus more on the style of image by filtering out most of its content information.
In this context, the style representation can be considered as a hardwired layer~\cite{3dcnn_icml2010} that encodes our prior knowledge.
We in particular adopt the Gram matrix of the feature map~\cite{gatys2016image} as the style representation as it provides richer style information than other methods, \eg, channel-wise feature statistics~\cite{Instancenorm}.
The Gram matrix, denoted by $\textbf{G} \in \mathbb{R}^{c \times c}$, captures correlations between $c$ channels of its input feature map.
The $(i,j)$ element of $\textbf{G}$ indicates the correlation between $i^\textrm{th}$ and $j^\textrm{th}$ feature channels and is computed by $\textbf{G}_{i,j} = \textbf{a}_{i}^\top \textbf{a}_{j}$,
where $\textbf{a}_{i}$ is the vector form of the $i^\textrm{th}$ channel of the input feature map.
Specifically, since the Gram matrix is symmetric, the vector form of only the upper triangular part of the matrix is used as input to the fog-pass filtering module. 

Let $I^a$ and $I^b$ be a pair of images from the mini-batch and $F^l$ denote the fog-pass filter attached to the $l^\textrm{th}$ layer of the segmentation network. 
Then the fog factors of the two images are computed by $\textbf{f}^{a,l}=F^l(\textbf{u}^{a,l})$ and $\textbf{f}^{b,l}=F^l(\textbf{u}^{b,l})$, respectively, where $\textbf{u}^{a,l}$ and $\textbf{u}^{b,l}$ denote the vectorized upper triangular parts of the Gram matrices computed from their $l^\textrm{th}$ intermediate feature maps.
The role of the fog-pass filter is to inform the segmentation network how $I^a$ and $I^b$ are different in terms of fog condition through $\textbf{f}^{a,l}$ and $\textbf{f}^{b,l}$.
For this purpose, the fog-pass filter learns a space of fog factors where those of the same fog domain are grouped closely together and those of different domains are far from each other. 
Given the set of every image pair $\mathcal{P}$ in the mini-batch, the loss function for $F^l$ is designed as follows:
\vspace{-1mm}
\begin{align} 
    \mathcal{L}_{F^l} = \sum_{(a,b)\in \mathcal{P}}\bigg\{\big(1-\mathbb{I}(a,b)\big) {\Big[m - d\big(\textbf{f}^{a,l}, \textbf{f}^{b,l} \big) \Big]_{+}^2} \nonumber \\ +  \mathbb{I}(a,b) {\Big[ d\big(\textbf{f}^{a,l}, \textbf{f}^{b,l}\big) - m \Big]_{+}^2}\bigg\},
    \label{eq:FFmodule_loss}
\vspace{-2mm}
\end{align}
where $d(\cdot)$ is the cosine distance, $m$ is a margin, and $\mathbb{I}(a,b)$ denotes the indicator function that returns 1 if $I^a$ and $I^b$ are of the same fog domain and 0 otherwise.

\begin{figure*}[t]
    \centering
    \vspace{-2mm}
    \includegraphics[width=0.95\linewidth]{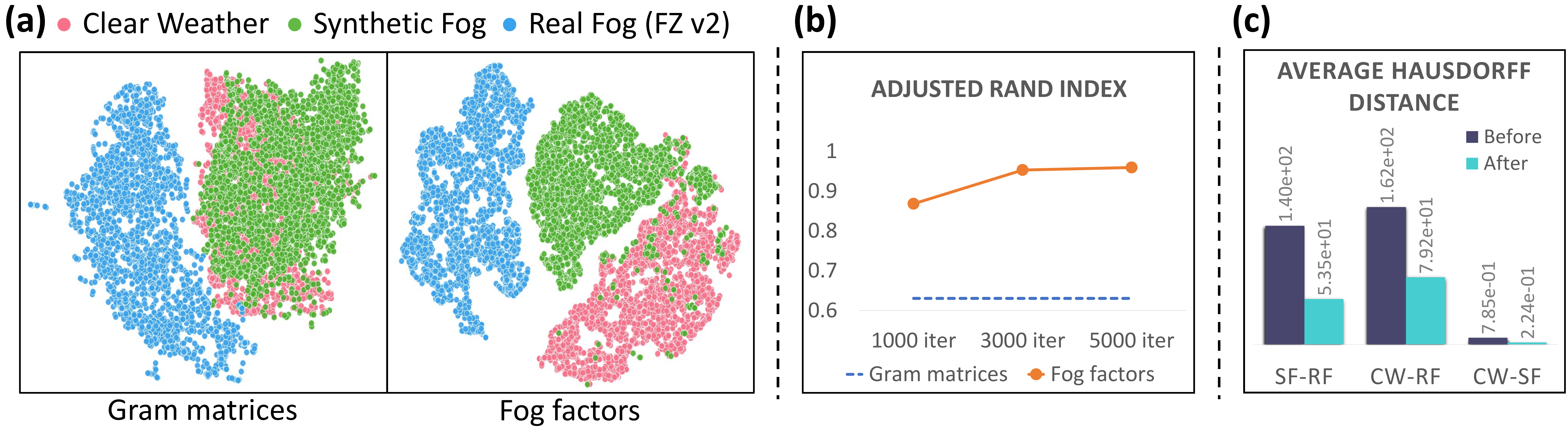}
    \vspace{-2mm}
\caption{   
Empirical analysis on the impact of FIFO. 
(a) 2D visualization of distributions of Gram matrices and their fog factors.
(b) Comparison between the quality of $k$-means clustering of the Gram matrices and that of the corresponding fog factors in adjusted Rand index.
(c) The fog-style gap between different domains before and after training with FIFO, where the gap is measured by the average Hausdorff distance between two sets of fog factors.
} \label{fig:TSNE}
\vspace{-3mm}
\end{figure*}

\subsection{Segmentation Network} \label{sec:segnet_detail}

The segmentation network is trained using three different objectives, which are designed for semantic segmentation, fog-invariant feature learning, and consistent prediction regardless of fog condition of input, respectively.
We elaborate on each of the loss functions below.

\noindent\textbf{Segmentation Loss.} 
For learning semantic segmentation, we apply the pixel-wise cross-entropy loss to individual images.
To be specific, the loss is given by
\begin{equation}
    \mathcal{L}_\textrm{seg}(\mathbf{P}, \mathbf{Y}) = -\frac{1}{n}\sum_{i}\sum_{j} \mathbf{Y}_{i,j}\log \mathbf{P}_{i,j},
    \label{eq:cross_entropy_loss}
\end{equation}
where $\mathbf{P}_{i,j}\in \mathbb{R}$ and $\mathbf{Y}_{i,j}\in \{0,1\}$ denote the predicted score and groundtruth label of class $j$ at pixel $i$, respectively, while $n$ is the number of pixels.

\noindent\textbf{Fog Style Matching Loss.}
Given a pair of images from different fog domains, the segmentation network learns fog-invariant features that close the distance between their fog factors. 
To this end, the second loss matches the two fog factors given by the frozen fog-pass filters.
Let $\textbf{f}_{i}^{a,l}$ and $\textbf{f}_{i}^{b,l}$ be the fog factors of the images computed by the fog-pass filter $F^l$.
Then the loss is given by 
\begin{equation} 
    \mathcal{L}^l_\textrm{fsm}(\textbf{f}^{a,l}, \textbf{f}^{b,l}) = {\frac{1}{4 d_l^2 n_l^2} \sum_{i=1}^{d_l} \Big(  \textbf{f}_{i}^{a,l} - \textbf{f}_{i}^{b,l} \Big)^2},
    \label{eq:fsm_loss}
\end{equation}
where $d_l$ and $n_l$ denote the dimension of their fog factors and the spatial size of the $l^\textrm{th}$ feature map, respectively.

\noindent\textbf{Prediction Consistency Loss.} 
A \sftype{CW} image and its \sftype{SF} counterpart
have exactly the same semantic layout. 
By forcing predictions for these images being identical, we can align the \sftype{CW} and \sftype{SF} domains more aggressively in the learned representation.
Hence, only for \sftype{CW} and \sftype{SF} images of the same origin, we encourage the model to predict the same segmentation map.
Let $\mathbf{P}_i^\sftypes{CW}\in \mathbb{R}^c$ and $\mathbf{P}_i^\sftypes{SF}\in \mathbb{R}^c$ denote their class probability vectors predicted by the segmentation model for pixel $i$, where $c$ is the number of classes. 
The third loss is designed to force the consistency between $\mathbf{P}_i^\sftypes{CW}$ and $\mathbf{P}_i^\sftypes{SF}$ for all pixels, and is given by
\begin{equation}
    \mathcal{L}_\textrm{con}(\mathbf{P}^\sftypes{CW}, \mathbf{P}^\sftypes{SF}) = \sum_{i}  \textrm{KLdiv}(\mathbf{P}_i^\sftypes{CW}, \mathbf{P}_i^\sftypes{SF}), \label{eq:kd_loss}
\end{equation}
where KLdiv$(\cdot,\cdot)$ is the Kullback–Leibler divergence.
This loss shares the same goal with the fog style matching loss in Eq.~(\ref{eq:fsm_loss}), but more strongly forces fog-invariance in the prediction level via a small number of \sftype{CW}--\sftype{SF} pairs.
Also, it is complementary to the segmentation loss in Eq.~(\ref{eq:cross_entropy_loss}) since the probability distributions in Eq.~(\ref{eq:kd_loss}) provide information beyond the class labels used by the segmentation loss.

\noindent\textbf{Training Strategy.}
Given a mini-batch, the same number of image pairs are sampled from each of the three different domain pairs,
\ie, \sftype{CW}--\sftype{SF}, \sftype{CW}--\sftype{RF}, and \sftype{SF}--\sftype{RF}.
Note that images of each \sftype{CW}--\sftype{SF} pair are of the same semantic layout so that the prediction consistency loss is applied to them. 
For \sftype{CW}--\sftype{SF} pairs, the segmentation network is trained by minimizing
\begin{align}
    & \mathcal{L}^{\sftypes{CW}\textrm{-}\sftypes{SF}}_S = \mathcal{L}_\textrm{seg}(\mathbf{P}^\sftypes{CW}, \mathbf{Y}^\sftypes{CW}) + \mathcal{L}_\textrm{seg}(\mathbf{P}^\sftypes{SF}, \mathbf{Y}^\sftypes{SF}) \nonumber \\ 
    &\phantom{00}+ \lambda_\textrm{fsm} \sum_{l} \mathcal{L}^l_\textrm{fsm}(\textbf{f}^{\sftypes{CW},l},\textbf{f}^{\sftypes{SF},l}) + {\lambda_\textrm{con}}\mathcal{L}_\textrm{con}(\mathbf{P}^\sftypes{CW},\mathbf{P}^\sftypes{SF}),
    \label{eq:clear_synthetic_objective}
\end{align}
where ${\lambda_\textrm{fsm}}$ and ${\lambda_\textrm{con}}$ are balancing hyper-parameters and $\mathbf{Y}^\sftypes{CW} = \mathbf{Y}^\sftypes{SF}$.
On the other hand, for the other pairs of input domains including \sftype{RF}, the loss consists of the segmentation and fog style matching terms only, and is given by
\begin{equation} 
    \mathcal{L}^{\sftypes{D}\textrm{-}\sftypes{RF}}_S = \mathcal{L}_\textrm{seg}(\mathbf{P}^\sftypes{D}, \mathbf{Y}^\sftypes{D}) + \lambda_\textrm{fsm} \sum_{l} \mathcal{L}^l_\textrm{fsm}(\textbf{f}^{\sftypes{D},l},\textbf{f}^{\sftypes{RF},l}),
    \label{eq:synthetic_real_objective}
\end{equation}
where $\sftype{D} \in \{ \sftype{CW}, \sftype{SF} \}$.
Note that $\mathcal{L}_\textrm{seg}$ is not applied to the prediction for real foggy image $\mathbf{P}^\sftypes{RF}$ due to the absence of its segmentation label.

\subsection{Empirical Verification}

\begin{figure*}
    \centering
    \vspace{-2mm}
    \includegraphics[width=0.92\textwidth]{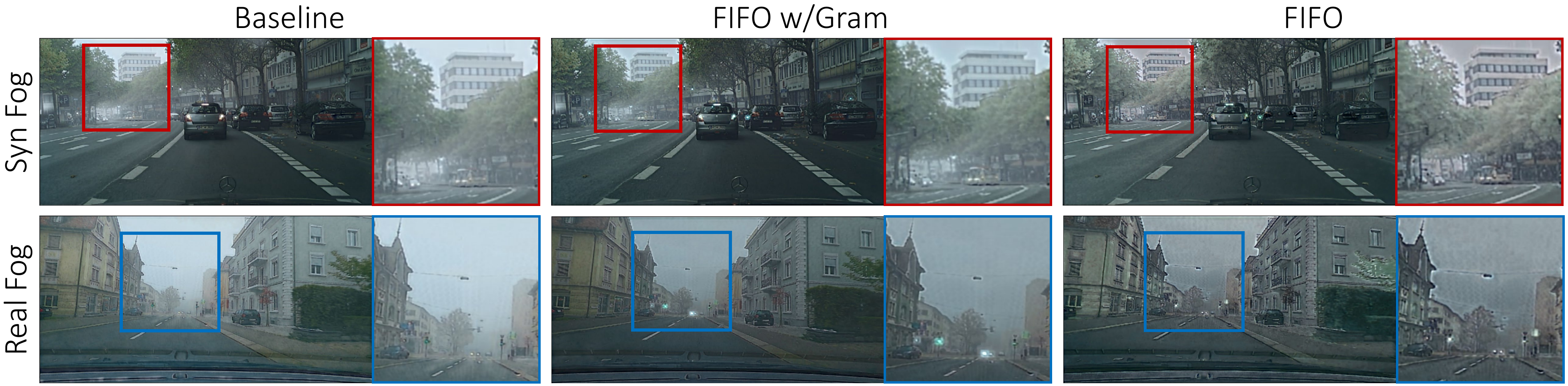}
    \vspace{-2.5mm}
    \caption{Images reconstructed by the baseline, a variant of FIFO closing the gap between Gram matrices, and FIFO.}
    \label{fig:Recon}
    \vspace{-3mm}
\end{figure*}

\noindent\textbf{Impact of Fog-pass Filtering Modules.} 
To justify the use of the fog-pass filters,  
we compare Gram matrices and their fog factors computed by a fog-pass filter in how well they disentangle the fog condition and the other aspects of an image.
To this end, we examine distributions of Gram matrices and fog factors of \sftype{CW}, \sftype{SF}, and \sftype{RF}.
To be specific, training images of the Cityscapes dataset~\cite{cityscapes}, their synthetic foggy counterparts given by the fog simulator~\cite{dai2020curriculum} with the attenuation coefficient $\beta=0.005$, and those of the Foggy Zurich dataset~\cite{Sakaridis_2018_ECCV} are adopted as \sftype{CW}, \sftype{SF}, and \sftype{RF} images, respectively;
their Gram matrices are computed from ResBlock1 outputs of RefineNet-lw~\cite{nekrasov2018light} pre-trained on the Cityscapes, and the corresponding fog factors are computed from the Gram matrices through the fog-pass filtering module.
Fig.~\ref{fig:TSNE}(a) presents $t$-SNE visualization~\cite{MaatenNov2008} of the distributions, in which Gram matrices of \sftype{CW} and \sftype{SF} largely overlap each other while fog factors are well separated according to their fog domains. 
This result suggests that Gram matrices are affected substantially by image content while fog factors represent only fog-relevant information as desired.
The same trend is observed in Fig.~\ref{fig:TSNE}(b) that quantitatively evaluates the quality of $k$-means clusters~\cite{hartigan1979algorithm} of the Gram matrices and fog factors via adjusted Rand index~\cite{hubert1985comparing}.

\noindent\textbf{Fog-invariance Learned by FIFO.} To investigate the impact of FIFO on fog-invariance learning, 
we first demonstrate that FIFO effectively reduces the gap between fog domains in the space of fog factors.
To this end, each domain is represented as the set of fog factors of its images, and the gap between a pair of domains is measured by the average Hausdorff distance~\cite{dubuisson1994modified} between such sets of the domains before and after training with FIFO.
Fig.~\ref{fig:TSNE}(c) shows that FIFO closes the fog-style gap in all three domain pairs.
The impact is also verified qualitatively through images reconstructed from intermediate features of the segmentation network trained by FIFO.\footnote{We conduct this experiment only for qualitative analysis on the effect of FIFO. Note that FIFO directly learns fog-invariant features while bypassing explicit fog removal of input image, thus does not reconstruct images at all in both training and testing.}
As a reconstruction model, we adopt RefineNet-lw with two additional upsampling layers;
its decoder is first trained to reconstruct images while freezing its encoder pretrained on the Cityscapes dataset, then the encoder is replaced with that of the segmentation network trained by FIFO.    
Also, for comparisons, we reconstruct images using a baseline (\ie, training only on the Cityscapes) and a variant of FIFO with no fog-pass filter (\ie, training by directly reducing the gap between Gram matrices) in the same manner.
Fig.~\ref{fig:Recon} presents examples of the reconstructed images, which demonstrate that FIFO allows to sharpen images effectively, well emphasize object boundaries in particular, and make fewer artifacts.

\section{Experiments} 

\subsection{Implementation Details} \label{implementation_details}

\noindent\textbf{Network Architecture.}
We adopt RefineNet-lw~\cite{nekrasov2018light} with ResNet-101~\cite{resnet} backbone as our segmentation network.
Two fog-pass filtering modules are then respectively attached to the outputs of Conv1 and ResBlock1 layers of the segmentation network.
As illustrated in Fig.~\ref{fig:detail}(a), the two modules are implemented by multi-layer perceptrons with leaky ReLU activation functions~\cite{leakyReLU}.

\noindent\textbf{Optimization and Hyper-parameters.}
The segmentation network is trained by SGD with a momentum of 0.9 and the initial learning rate of $\expnum{6}{4}$ for the encoder and $\expnum{6}{3}$ for the decoder; both learning rates are decreased by polynomial decay with a power of 0.5. 
The two fog-pass filtering modules are trained by Adamax~\cite{Adamsolver} with initial learning rates of $\expnum{5}{4}$ (Conv1) and $\expnum{1}{3}$ (ResBlock1), respectively;
the dimensionality of fog factors is set to 64.
Each mini-batch is constructed by sampling 4 images from each fog domain, thus its size is 12. 
During training, images are resized, cropped to $600 \times 600$, and flipped horizontally at random.
Finally, the hyper-parameters $\lambda_\textrm{fsm}$, $\lambda_\textrm{con}$, and $m$ are set to $\expnum{5}{8}$, $\expnum{1}{4}$, and 0.1, respectively.

\subsection{Datasets for Evaluation} \label{subsec:evaldataset}

FIFO is evaluated and compared with previous work on three real foggy datasets: {Foggy Zurich (FZ) test v2}~\cite{Sakaridis_2018_ECCV}, {Foggy Driving (FD)}~\cite{Sakaridis_2018_IJCV}, and {Foggy Driving Dense (FDD)}~\cite{Sakaridis_2018_ECCV}; FDD is a subset of FD.
Images of these datasets depict various fog densities and are fully annotated.
Also, they share the same class set with the Cityscapes dataset as described in~\cite{dai2020curriculum}.
We further apply FIFO and previous work to an unseen clear weather dataset, Cityscapes lindau 40 introduced in~\cite{dai2020curriculum}, to evaluate their performance on clear weather scenes as well. 

\begin{table*}[!htb]
\setlength\tabcolsep{0pt}
\small
\smallskip
\renewcommand{\arraystretch}{1.05}
\centering
\vspace{-2mm}
\begin{tabular*}{2\columnwidth}{@{\extracolsep{\fill}}l|cc|ccc|c}
\toprule
\multirow{2}{*}{Method} & \begin{tabular}[c]{@{}c@{}}Clear-weather\end{tabular} & Synthetic ~\&~ Real fog~~~ & FZ test v2~\cite{Sakaridis_2018_ECCV}    & FDD~\cite{Sakaridis_2018_ECCV}   & ~~FD~\cite{Sakaridis_2018_IJCV}~~ & Cityscapes lindau 40~\cite{cityscapes}      \\
& Cityscapes~\cite{cityscapes}                                            &  ~~SDBF~\cite{Sakaridis_2018_ECCV}~~ GoPro~\cite{Sakaridis_2018_ECCV}~~~~& mIoU (\%) & mIoU (\%) & ~~~ mIoU (\%)~~~~ & mIoU (\%) \\ \hline
RefineNet~\cite{lin2017refinenet}             & \checkmark  &             &  34.6      &  35.8      &  44.3   &  67.2             \\
RefineNet-lw~\cite{nekrasov2018light} ~       & \checkmark  &             &  28.5      &  35.9      &  43.6   &  63.8             \\
\hline
AdSegNet~\cite{AdaptSegNet}          & \checkmark  & \checkmark  &  25.0      &  15.8      &  29.7   &  -             \\
AdvEnt~\cite{vu2019advent}              & \checkmark  & \checkmark &  39.7      &  41.7      &  46.9   &  61.7             \\
FDA~\cite{yang2020fda}             & \checkmark  & \checkmark &  22.2   &  29.8   &  21.8      &  39.3            \\
DANN~\cite{dann}      & \checkmark  & \checkmark & 43.1    & 41.4    & 46.0       & 60.1             \\
DANN-Gram       & \checkmark  & \checkmark  & 43.4    & 43.3    & 47.3       & 67.1             \\
\hline
CMAda2+~\cite{dai2020curriculum}             & \checkmark  & \checkmark  &  43.4      &  40.1      &  49.9   &  -             \\
CMAda3+~\cite{dai2020curriculum}             & \checkmark  & \checkmark  &  46.8      &  43.0      &  49.8   &  59.6             \\ 
FIFO                   & \checkmark  & \checkmark  & \textbf{48.4} & \textbf{48.9} & \textbf{50.7}  &  64.8 \\  
\bottomrule
\end{tabular*}
\vspace{-2mm}
\caption{
    Quantitative results in mean intersection over union (mIoU) on three real foggy datasets---Foggy Zurich (FZ) test v2, Foggy Driving Dense (FDD), Foggy Driving (FD), and a clear weather dataset---Cityscapes lindau 40. 
}
\vspace{-1mm}
\label{tab:main_table}
\end{table*}

\begin{figure*}[t]
    \centering
    \vspace{-2mm}
    \includegraphics[width=0.99\linewidth]{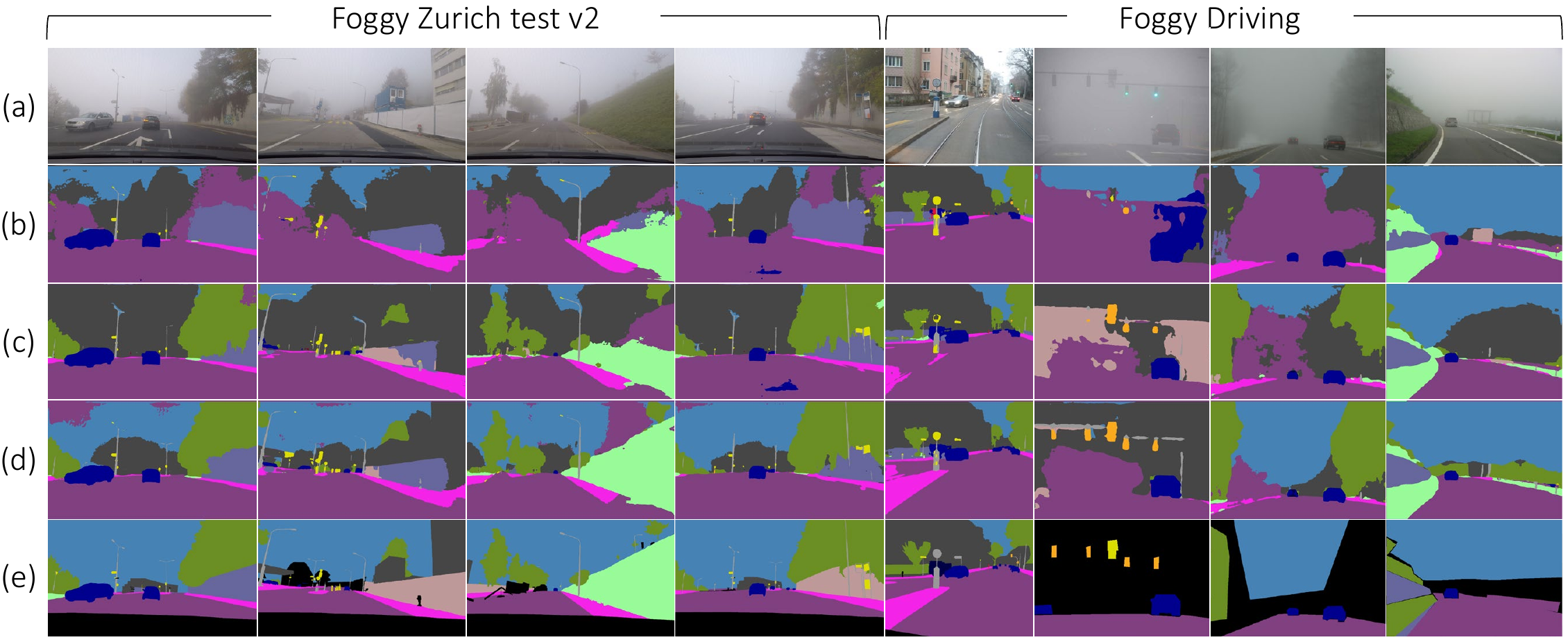}
\vspace{-3mm}
\caption{Qualitative results on the real foggy datasets.
(a) Input images. 
(b) Baseline.
(c) FIFO without the fog-pass filtering. (d) FIFO. (e) Groundtruth.
} 
\label{fig:qualitativeresult}
\vspace{-3mm}
\end{figure*}

\subsection{Quantitative Analysis} \label{subsec:performance}
Quantitative results of FIFO and previous arts are summarized in Table~\ref{tab:main_table}.
As shown in the table, FIFO largely outperforms CMAda3+~\cite{dai2020curriculum}, the current best performing model based on RefineNet backbone~\cite{lin2017refinenet}, on all the three foggy image datasets.
These results indicate that our method of closing the fog style gap is superior to the curriculum adaptation on real images with various fog densities.

We also validate the performance of the models on the clear weather dataset. 
In this experiment, the accuracy of CMAda3+ drops substantially; we suspect this is a side effect of the curriculum adaptation, which may lead to overfitting to foggy scenes due to catastrophic forgetting.
On the other hand, FIFO enhances performance on clear weather, probably because of the data augmentation effects of using the three domains altogether during training.

\subsection{Qualitative Results}
\label{subsec:qualitative}
FIFO is qualitatively compared with two other methods.
One is RefineNet-lw trained only on the Cityscapes dataset, which we call baseline. 
The other is a reduced version of FIFO using no fog-pass filtering module; this method learns the segmentation network while closing the gap between Gram matrices from different domains. 
Qualitative examples of their predictions are presented in Fig.~\ref{fig:qualitativeresult}.
The baseline yields poor results in most cases.
The reduced version of FIFO outperforms the baseline, especially for car, road, and vegetation classes, but FIFO using the fog-pass filtering modules clearly demonstrates the best segmentation results.

\subsection{Comparison with UDA}
The task of FIFO is ostensibly identical to that of unsupervised domain adaptation (UDA) since both of them adapt models to an unlabeled target domain.
Hence, one may wonder how well UDA models work for semantic foggy scene segmentation under the setting of FIFO. 
In this context, FIFO is compared with multiple UDA methods based on various levels of adaptation: FDA~\cite{yang2020fda} for input-level adaptation, AdSegNet~\cite{AdaptSegNet} and AdvEnt~\cite{vu2019advent} for output-level adaptation, and DANN~\cite{dann} for feature-level adaptation.
We also evaluate a variant of DANN whose domain classifier takes as input a Gram matrix, instead of a raw feature map, like FIFO; this variant is denoted by DANN-Gram. 
The UDA models are trained using the same datasets, \ie, \sftype{CW}, \sftype{SF}, and \sftype{RF}.
The input- and output-level adaptation models are pretrained on \sftype{CW} and further trained using \sftype{SF} while adapting to \sftype{RF}.
DANN and DANN-Gram are trained using \sftype{CW}, \sftype{SF}, and \sftype{RF} at once like FIFO: 
Their discriminators are optimized to maximize the discrepancy of fog domains like our fog-pass filters while their segmentation networks are learned to minimize the discrepancy.

The performance of these UDA methods is reported in Table~\ref{tab:main_table}.
As shown in the table, the UDA models are all inferior to FIFO and CMAda3+, which suggests that, even though it is apparently similar to UDA for semantic segmentation, semantic foggy scene segmentation is of a different nature and has its own challenges. 
In the typical UDA setting, each source and target domain has its own style, by which it can be defined.
However, the style of a foggy scene is not determined only by its fog condition, rather is the result of the chemical combination of fog and other style factors of the scene.
The UDA methods that consider \sftype{SF} and \sftype{RF} as domains with unique styles are thus not well suited to the task. 
Moreover, fog substantially damages visibility, and enlarges intra-domain variations since the effect of fog varies significantly according to the 3D configuration of the scene~\cite{dai2020curriculum}.
Due to these differences and challenges, foggy scene segmentation demands dedicated solutions like FIFO.

\begin{table}[t!]
  \centering
  \vspace{-2mm}
    \small
    \renewcommand{\arraystretch}{1.12}
    \scalebox{1}{\begin{tabular*}{1\columnwidth}{ccc|cccc}
    \toprule
\sftype{CW}--\sftype{SF}                   & \sftype{CW}--\sftype{RF}                & \sftype{SF}--\sftype{RF}                & FZ     & FDD & FD & CW\\ \hline
           \checkmark    &                         &                         &  43.7         & 38.6  & 46.1    & 67.6 \\
                             & \checkmark &                         &  37.7        & 40.3  & 47.2    & 66.0 \\
    &                         & \checkmark & 39.3          & 42.8  & 49.7    & 61.6    \\
    \hline
    \checkmark    & \checkmark &                         & 49.7          & 46.0  & 49.9    & 65.8     \\
    \checkmark    &                         & \checkmark & 46.0          & 47.6  & 50.0    & 62.3  \\
     & \checkmark & \checkmark &47.4          & 38.2  & 47.0    & 64.3      \\
    \hline
      \checkmark    & \checkmark & \checkmark & 48.4 & 48.9 & 50.7  &  64.8 \\ 
    \bottomrule
    \end{tabular*}}
    \vspace{-2mm}
    \caption{
    Analysis on the impact of domain pairs. \sftype{CW}, \sftype{SF} and \sftype{RF} denote clear weather, synthetic fog, and real fog, respectively.
  }
    \label{tab:ablation_domain_pair}
    \vspace{-2mm}
\end{table}

\begin{table}[t!]
\vspace{-1mm}
  \centering
  \scalebox{0.95}{\begin{minipage}{.54\linewidth}
    \centering
    \renewcommand{\arraystretch}{1.2}
    \scalebox{0.8}{\begin{tabular*}{1.27\columnwidth}{lccc}
    \toprule
    Method                     & FZ  & FDD  & FD    \\ \midrule
    Baseline                   &  28.5           & 35.9          & 43.6                 \\
    FIFO w/o $\mathcal{L}_\textrm{fsm}$      & 31.7	         & 38.5	         & 45.1	               \\
    FIFO w/o $\mathcal{L}_\textrm{con}$           & 41.6            & 45.4          & 48.9                 \\
    FIFO w/ Gram                      & 41.3 & 	43.8 &	49.1   \\ \midrule
    FIFO                       & \textbf{48.4}   & \textbf{48.9} & \textbf{50.7}        \\ \bottomrule
    \end{tabular*}}
    \vspace{-2mm}
    \caption{
        Analysis on the impact of the fog style matching loss, the prediction consistency loss, and the fog-pass filtering modules.
     }    
    \label{tab:ablation_losses}
  \end{minipage}}
  \quad 
  \scalebox{0.95}{\begin{minipage}{0.45\linewidth}
        \centering
      \vspace{0mm}
      \renewcommand{\arraystretch}{1.2}
        \scalebox{0.8}{
        \begin{tabular*}{1.2\columnwidth}{lccc}
        \toprule
        $\beta$          & FZ           & FDD             & FD                   \\ \midrule
        0                & 37.7                 & 40.3            & 47.2                \\
        0.0025           & 45.5                 & 40.4            & 45.0            \\
        
        0.01             & 42.4                 & 45.4            & 50.0             \\
        0.02             & 42.9                 & 42.5            & 48.6                \\ \midrule
        0.005             & \textbf{48.4}   & \textbf{48.9} & \textbf{50.7}                     \\\bottomrule
        \end{tabular*}
        }
        \vspace{-2mm}
    \caption{
    Analysis on the impact of synthetic fog densities.
    $\beta=0$ denotes a model trained with \sftype{CW} and \sftype{RF}.
    }
    \label{tab:beta_ablation}
    \vspace{0mm}
  \end{minipage}} 
  \vspace{-3mm}

\end{table}

\subsection{Ablation Study} \label{subsec:ablation}

We conduct extensive experiments while varying domain pairs of FIFO to investigate their effects.
Table~\ref{tab:ablation_domain_pair} summarizes the results.
We found that using more pairs in general boosts performance, which suggests that all the three pairs contribute to performance on real foggy images.

We also investigate contributions of the fog style matching loss $\mathcal{L}_\textrm{fsm}$, the prediction consistency loss $\mathcal{L}_\textrm{con}$, and the fog-pass filtering modules to the performance. 
Table~\ref{tab:ablation_losses} compares FIFO with its variants with and without $\mathcal{L}_\textrm{fsm}$, $\mathcal{L}_\textrm{con}$, and the fog-pass filters in terms of segmentation quality on real foggy images.
Note that FIFO w/o $\mathcal{L}_\textrm{fsm}$ is trained by the segmentation and prediction consistency losses only, dropping the fog style matching loss, while FIFO w/ Gram uses Gram matrices instead of fog factors when matching fog styles. 
The results in the tables suggest that all the losses and the fog-pass filtering contribute to the performance on all the three real foggy datasets, 
but the impact of the fog style matching is substantially larger than the others.
Also, the gap between FIFO and FIFO w/ Gram demonstrates the superiority of fog factors over Gram matrices, which justifies the use of the fog-pass filters.

In addition, we demonstrate the effect of our chosen value of $\beta$, which is the attenuation coefficient used for generating synthetic fog~\cite{Sakaridis_2018_IJCV}.
Note that our method exploits a single value of $\beta$ to resolve the issue of the curriculum adaptation~\cite{Sakaridis_2018_ECCV,dai2020curriculum}.
Table~\ref{tab:beta_ablation} summarizes the performance of variants of FIFO trained using different values of $\beta$;
the optimal value for $\beta$ is 0.005.

Finally, we examine the impact of the layers to which the fog style matching loss $\mathcal{L}_{fsm}$ is applied.
Specifically, $\mathcal{L}_{fsm}$ is applied to the output of the first convolutional layer (C1), the first residual block (R1), or both of them (C1+R1).
As summarized in Table~\ref{tab:layerablation}, the performance improves as more feature maps are affected by the loss.
Overall, C1+R1, which is our final model, shows the best performance.

\subsection{Generalization to Other Weather Conditions} \label{subsec:generalization}
\begin{figure}
    \centering
    \vspace{-1mm}
    \includegraphics[width=0.47\textwidth]{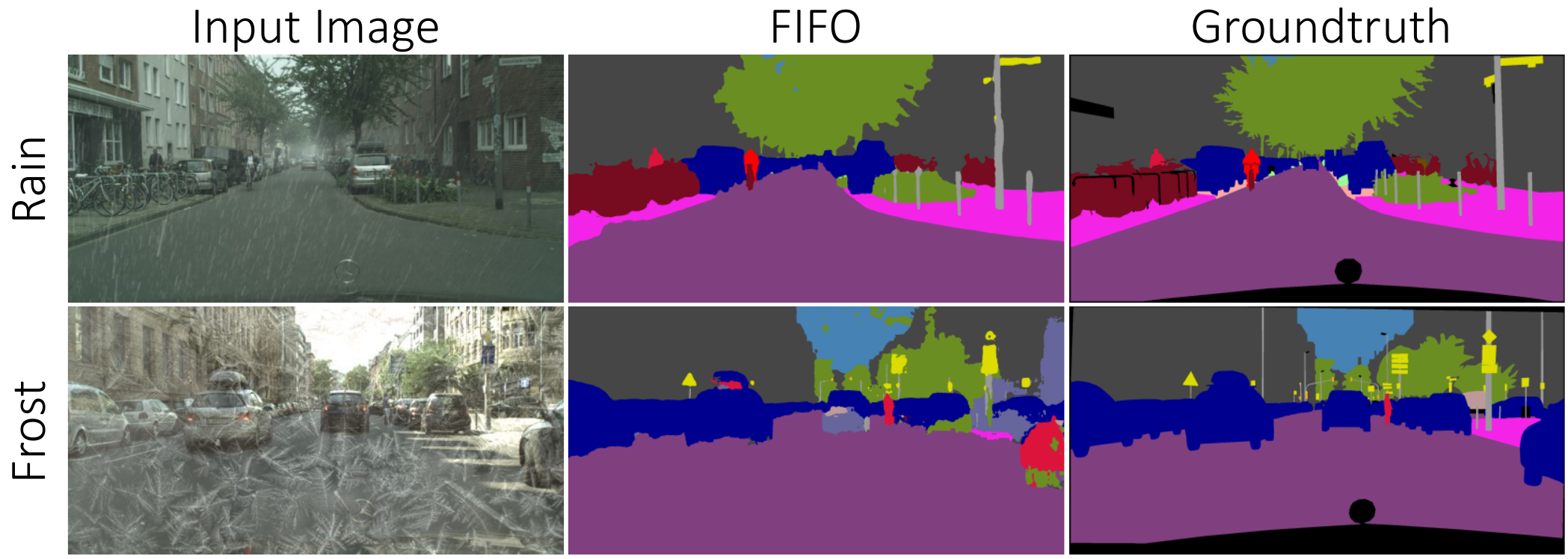}
    \vspace{-3mm}
    \caption{
    Qualitative results under rain and frost conditions.
    }
    \label{fig:weather}
    \vspace{-3mm}
\end{figure}

We investigate the generalization ability of FIFO on other weather conditions, \emph{rain} and \emph{frost}.
To this end, RefineNet-lw trained by FIFO in Sec.~\ref{sec:method} is evaluated \emph{as-is} on rainy~\cite{Hu_2019_CVPR} and frosty~\cite{Hendrycks2019_ImageNetC} versions of the Cityscapes dataset.
On the rainy Cityscapes dataset, our model is compared with existing methods reported in~\cite{franchi} that aim at improving robustness using \sftype{CW} images only.
As summarized in Table~\ref{tab:city_rainy}, our model largely outperforms the previous work based on a stronger segmentation network (\ie, DeepLab v3+~\cite{chen2018encoder}, indicated by $\dagger$ in the table).
Fig.~\ref{fig:weather} demonstrates that FIFO generalizes to the frosty images also.
More results can be found in the supplementary material.

\begin{table}[t!]
\vspace{-3mm}
  \centering
  \scalebox{0.9}{\begin{minipage}{.61\linewidth}
    \centering
    \small
    \renewcommand{\arraystretch}{1.0}
    \scalebox{1}{\begin{tabular*}{1.0\columnwidth}{lccc}
    \toprule
               & FZ  & FDD  & FD       \\ \midrule
       
    C1                  & 45.3              & 41.0            & 48.3                     \\
    R1                &   45.1            &  39.1          &   47.0                   \\
    C1+R1 (Ours)       & \textbf{48.4}               & \textbf{48.9}           & \textbf{50.7}                 \\\bottomrule
    \end{tabular*}}
    \vspace{-2mm}
    \caption{
    Analysis on the layers to which the fog style matching loss is applied. C1 and R1 indicate the output of the first convolution layer and that of the first residual block, respectively.
    }
    \vspace{-0.8mm}
    \label{tab:layerablation}
  \end{minipage}}
  \quad 
  \scalebox{0.9}{\begin{minipage}{0.44\linewidth}
        \centering
      \vspace{1mm}
        \scalebox{0.88}{\begin{tabular*}{1.18\columnwidth}{lc}
        \toprule
        Method                     & RC     \\ \midrule
        $^\dagger$Baseline~\cite{hendrycks2016baseline}                   &  59.0                \\
        $^\dagger$Cutmix~\cite{french2019semi}              &  61.9             \\ 
        $^\dagger$LP-BNN~\cite{franchi2020encoding}              &  60.7             \\ 
        $^\dagger$Superpixel-mix~\cite{franchi}              &  61.9             \\ \midrule
        Baseline~\cite{nekrasov2018light}                   &  57.6              \\
        FIFO                       & \textbf{67.6}    \\ \bottomrule
        \end{tabular*}}
        \vspace{-2mm}
        \caption{
            Quantitative results on Rainy Cityscapes (RC).
        }    
        \label{tab:city_rainy}
    \vspace{0mm}
  \end{minipage}} 
  \vspace{-2mm}

\end{table}
\section{Conclusion} \label{sec:conclusion}

We have presented FIFO, a new approach to learning fog-invariant features for foggy scene segmentation.
It precisely quantifies the fog style of an image through the fog-pass filtering modules and learns a segmentation network for closing the gap between images of different fog conditions in the fog style space.
FIFO outperforms previous arts without sacrificing performance on clear weather images. 
Moreover, unlike the current best-performing method, it enables end-to-end learning and demands no extra module nor human intervention for training.

\vspace{2mm}
{\small
\noindent \textbf{Acknowledgement:} This work was supported by Samsung Research Funding \& Incubation Center of Samsung Electronics under Project Number SRFC-IT1801-05.
}

\setcounter{section}{0}
\def\thesection{\Alph{section}}
\section*{\Large{Appendix}}
\renewcommand\thefigure{A\arabic{figure}}
\renewcommand{\thetable}{A\arabic{table}}
\setcounter{figure}{0}
\setcounter{table}{0}

\section{Algorithm of FIFO} \label{sec:algorithm}
We present the training procedure of FIFO in Algorithm~\ref{alg:alg}.

\begin{algorithm} [H]
    \caption{: Training FIFO}\label{alg:alg}
    
    \textbf{Input:} Pretrained fog-pass filtering module for the $l^\textrm{th}$ layer: $F^l(\cdot)$, Segmentation network: $S(\cdot)$, Number of layers: $L$, Batch size per domain: $m$, Segmentation prediction: $P$, Segmentation label: $Y$, Input image set $\{I^\sftype{CW},I^\sftype{SF},I^\sftype{RF}\}$: $x$, Subset of two elements from domain set \{\sftype{CW},\sftype{SF},\sftype{RF}\}: $\{a,b\}$ and Segmentation label set $\{Y^\sftype{CW},Y^\sftype{RF}\}$: $y$.
    
    \textbf{Output:} Optimized segmentation network $S(\cdot)$.
    
    \begin{algorithmic}[1]
    
    \For{\{1, \dots , \# of training iterations\}}
        \State Sample mini-batch $\{ x_i \}_{i=1}^m$
        \For{\{{$l \longleftarrow 1$ to $L$}\}}
        \State $\mathcal{L}_{\textrm{F}^l} \longleftarrow \mathcal{L}_{\textrm{F}^l}(\{\textbf{f}^l_i\}_{i=1}^m)$ 
        \State Update the fog-pass filtering module $F^l$
        \EndFor
        \State Sample mini-batch $\{ x_j \}_{j=1}^m$ and $\{ y_j \}_{j=1}^m$ 
        \State Sample the pair $\{I^a,I^b\} \in x_j$
        \For{\{{$l \longleftarrow 1$ to $L$}\}}
        \State $\{\textbf{f}_j^{a,l}\}_{j=1}^m, \longleftarrow \{{F^l}(\textbf{u}_j^{a,l})\}_{j=1}^m$
        \State $\{\textbf{f}_j^{b,l}\}_{j=1}^m, \longleftarrow \{{F^l}(\textbf{u}_j^{b,l})\}_{j=1}^m$
        \State $\mathcal{L}^l_\textrm{fsm} \longleftarrow \{\mathcal{L}^l_\textrm{fsm}(\textbf{f}_j^{a,l}, \textbf{f}_j^{b,l})\}_{j=1}^m$
        \EndFor
        \If{$\{a,b\}==\{\sftype{CW},\sftype{SF}\}$}
        \State $\mathcal{L}_\textrm{con} \longleftarrow \sum_{i}  \textrm{KLdiv}(P_i^a, P_i^b)$ 
        \EndIf
        \If{$\{a,b\} \cap \{\sftype{CW},\sftype{SF}\} \neq \emptyset$}
        \State $\mathcal{L}_\textrm{seg} \longleftarrow -\frac{1}{n}\sum Y\log P$ 
        \EndIf
        \State $\mathcal{L}_{\textrm{S}} \longleftarrow \sum_l \mathcal{L}_{\textrm{fsm}}^l + \mathcal{L}_\textrm{con} + \mathcal{L}_\textrm{seg}$
        \State Update the segmentation network $S$
    \EndFor
    
    \end{algorithmic}
\end{algorithm}

Consequently, the total objective of FIFO is following: 
\begin{equation} 
\sum_l\min_{F^l} \mathcal{L}^l_{F^l} + \min_{S} (\sum_l \mathcal{L}_{\textrm{fsm}}^l + \mathcal{L}_\textrm{con} + \mathcal{L}_\textrm{seg}),
\label{eq:total_objective}
\end{equation}
\vspace{-2mm}
where $l$ is the layer index.

\vspace{2mm}
\section{Generalization to Other Weather Conditions} \label{sec:other_weather}
\begin{figure}[h]{}
    \centering
    \vspace{-2mm}
    \includegraphics[width=1\columnwidth]{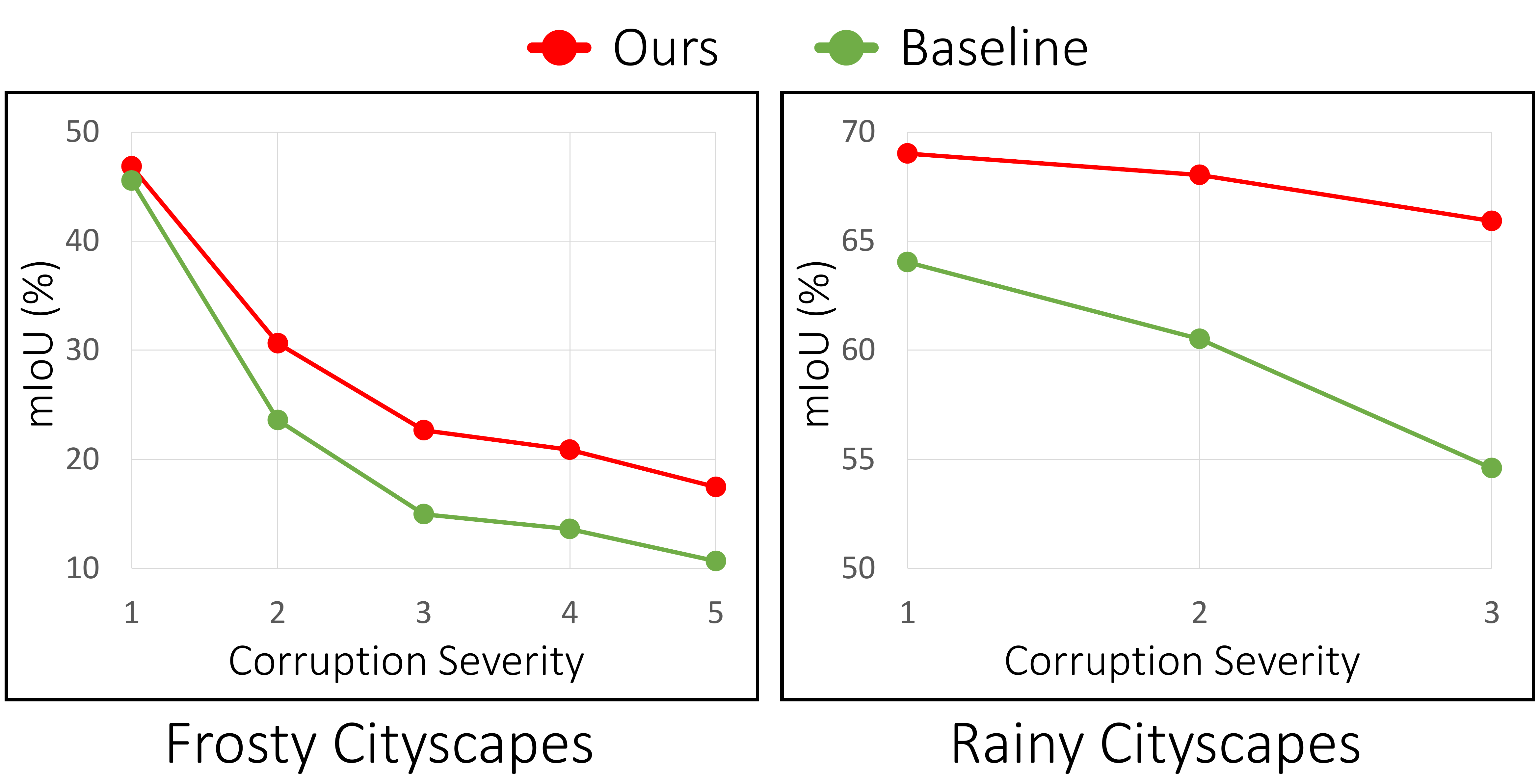}
\vspace{-6mm}
\caption{Performance (mIoU) versus the corruption severity.
Ours (FIFO) and baseline are evaluated on Frosty Cityscapes and Rainy Cityscapes.} 
\label{fig:other_weather}
\vspace{-3mm}

\end{figure}

We investigate the generalization ability of FIFO on the other weather conditions, \textit{rainy}~\cite{Hu_2019_CVPR} and \textit{frosty}~\cite{Hendrycks2019_ImageNetC} versions of the Cityscapes~\cite{cityscapes} dataset, according to the severity of the corruptions.
Figure~\ref{fig:other_weather} presents the performance of baseline~\cite{nekrasov2018light}, an ordinary segmentation model trained on clear weather images, and FIFO on varying the severity of frosty and rainy corruptions.
FIFO tends to be robust to each corruption than the baseline, even when the corruption gets severe. 
Table~\ref{tab:exp_frost} and Table~\ref{tab:exp_rain} show detailed quantitative results of baseline and FIFO on frosty and rainy corruptions, respectively.
Additional qualitative results are presented in Figure~\ref{fig:weather_qual}.

We also evaluate FIFO on ACDC, the real-world adverse conditions dataset for semantic driving scene understanding.
For fair comparisons on ACDC~\cite{acdc}, FIFO is trained on the Cityscapes, Foggy
Cityscapes-DBF, and Foggy Zurich datasets, following the unsupervised learning setting of the benchmark. As summarized in Table.~\ref{tab:ACDC}, FIFO outperforms the existing foggy scene segmentation methods reported in \cite{acdc} for all four conditions.

\begin{table}[h]
  \centering
    \centering
    \vspace{-2mm}
    \renewcommand{\arraystretch}{1.1}
    \scalebox{0.97}{
    \begin{tabular*}{1\columnwidth}{@{}C{1.3cm}@{}@{}C{1.5cm}@{}|@{}C{1.1cm}@{}@{}C{1.1cm}@{}@{}C{1.1cm}@{}@{}C{1.1cm}@{}@{}C{1.1cm}@{}}
    \toprule
                         &   & \multicolumn{5}{c}{Corruption Severity}     \\ 
                       &             & 1          & 2          & 3 & 4 & 5                  \\ \hline
    \multicolumn{1}{c|}{\multirow{2}{*}{Frost}} & Baseline & 45.53 & 23.59 & 14.97 & 13.60 & 10.66 \\
    \multicolumn{1}{c|}{\multirow{2}{*}{}} & FIFO & 46.85 & 30.64 & 22.66 & 20.88 & 17.47
    \\ \bottomrule
    \end{tabular*}
    }
    \vspace{-2mm}
    \caption{
        Quantitative results on Frosty Cityscapes according to the severity of corruptions.
  }    
    \label{tab:exp_frost}
    \vspace{-3mm}

\end{table}

\begin{table}[h]
  \centering
    \centering
    \renewcommand{\arraystretch}{1.1}
    \scalebox{0.95}{
    \begin{tabular*}{0.88\columnwidth}{@{}C{1.3cm}@{}@{}C{1.5cm}@{}|@{}C{1.1cm}@{}@{}C{1.1cm}@{}@{}C{1.1cm}@{}@{}C{1.1cm}@{}}
    \toprule
                         &   & \multicolumn{4}{c}{Corruption Severity}     \\ 
                       &             & 1          & 2          & 3     & all   \\ \hline
    \multicolumn{1}{c|}{\multirow{2}{*}{Rain}} & Baseline & 64.03 & 60.51 & 54.62 & 57.60 \\ 
    \multicolumn{1}{c|}{\multirow{2}{*}{}} & FIFO & 69.01 & 68.03 & 65.92 & 67.62 \\  \bottomrule
    \end{tabular*}
    }
    \vspace{-2mm}
    \caption{
        Quantitative results on Rainy Cityscapes according to the severity of corruptions.
  }    
    \label{tab:exp_rain}
    \vspace{-3mm}

\end{table}
\begin{table}[ht]
\vspace{0mm}
  \centering
  \renewcommand{\arraystretch}{1.1}
    \centering
    \vspace{-2mm}
    \normalsize
    \renewcommand{\arraystretch}{1.3}
    \scalebox{0.9}{\begin{tabular*}{0.9\columnwidth}{lccccc}
    \toprule
                Method         & Fog          & Rain             & Snow         &Night   &Avg.            \\ \midrule
              RefineNet         &    46.4        & 52.6         &  43.3     &     29.0      & 43.7     \\
              SFSU [2]        &    45.6        &  51.6        &  41.4      &    29.5       &   42.9   \\
              CMAda         &   51.2         &    53.4      &  47.6      &  32.0    &  47.1       \\
              \rowcolor[gray]{.8} 
              FIFO      &   \textbf{54.1}             &	\textbf{58.8}         &  \textbf{51.8}     &  \textbf{32.5}  &   \textbf{49.4}       \\\bottomrule
    \end{tabular*}}
    \vspace{-2mm}
    \caption{Quantitative results on the ACDC dataset.}
    \vspace{-5mm}
    \label{tab:ACDC}

\end{table}

\section{Independence Analysis of Fog Factors} \label{sec:independence}
In this section, we quantitatively evaluate the independence of the fog factors from the image content compared to that of the Gram matrices from the content.
To this end, we design a content-pass filtering module that is optimized to extract content-relevant information, which we call content factors.

\noindent\textbf{Training Content-pass Filtering Module.} 
Let $I^a$ and $I^b$ be a pair of images from the mini-batch, and $C^l$ denote the content-pass filtering module attached to the $l^\textrm{th}$ layer of the segmentation network.
Let $\textbf{u}^{a,l}$ and $\textbf{u}^{b,l}$ be the vectorized upper triangular parts of the Gram matrices computed from the $l^\textrm{th}$ feature maps of $I^a$ and $I^b$. 
Then the content factors of the two images are computed by  $\textbf{c}^{a,l}=C^l(\textbf{u}^{a,l})$ and $\textbf{c}^{b,l}=C^l(\textbf{u}^{b,l})$.
In contrast to the fog-pass filtering module, this module is optimized to learn an embedding space of content factors where the pairs having the same content, \ie, \sftype{CW}--\sftype{SF} are grouped closely and else pairs are far from each other.
Given the set of every image pair $\mathcal{P}$ in the mini-batch, the loss function for $C^l$ is designed accordingly as follows:
\begin{align}
\begin{split}
    \mathcal{L}_{C^l} = \sum_{(a,b)\in \mathcal{P}}\bigg\{\big(1-\mathbb{I}(a,b)\big) {\Big[m - d\big(\textbf{f}^{a,l}, \textbf{f}^{b,l} \big) \Big]_{+}^2} \\
    + \mathbb{I}(a,b) {\Big[ d\big(\textbf{f}^{a,l}, \textbf{f}^{b,l}\big) - m \Big]_{+}^2}\bigg\},
    \label{eq:contentfactor_module_loss}
    \end{split}
\end{align}
where $d(\cdot)$ is the cosine distance, $m$ is a margin, and $\mathbb{I}(a,b)$ denotes the indicator function that returns 1 if the pair of $I^a$ and $I^b$ is a \sftype{CW}--\sftype{SF} pair and 0 otherwise, respectively.

\noindent\textbf{Independence Analysis of Fog Factors.} 
We design the independence score to quantitatively evaluate and compare the independence of fog factors and that of Gram matrices from content factors.
We first measure the score of the independence of fog factors from content factors.
To this end, we select one image $I_i$, then choose k images $\{I_n\}$ whose fog factors are most similar to the fog factor $f_i$ of the selected image $I_i$.
Then, we also choose k images $\{I_m\}$ whose content factors are most similar to the content factor $c_i$ of the selected image $I_i$.
After that, we compute the proportion of the number of overlapped images $|\{I_n\} \cap \{I_m\}|$ between $\{I_n\}$ and $\{I_m\}$.
Then, we repeat the process for all $N$ images and calculate the average proportion as the independence score.

Let $I$, $f$, and $c$ be an image, a fog factor, and a content factor, then, the independence score is calculated as follows:
\begin{flalign}
    {\mathrm{Independence Score}} (\mathcal{F},\mathcal{C}) = \nonumber && 
\end{flalign}
\vspace{-6mm}
\begin{align} 
\begin{split}    
    1 - \frac{1}{N}\sum_{i=1}^N \frac{1}{k}\bigg\{\big|\{I_n | f_n \in \mathcal{F}, d(f_i,f_n) \leq d(f_i,f_k)\} \\ 
    \cap \{I_m | c_m \in \mathcal{C}, d(c_i,c_m) \leq d(c_i,c_k)\}\big|\bigg\},
    \label{eq:contentfactor_independence_score}
    \end{split}
\end{align}
where $d(\cdot)$ and $k$ are a cosine distance and a number of selecting similar factors set to 200, $f_k$ and $c_k$ are the $k$ th most similar fog factor from $f_i$ and the $k$ th most similar content factor from $c_i$, where $\mathcal{F}$ and $\mathcal{C}$ denote the set of fog factors and content factors, respectively.
We then replace the fog factors with Gram matrices, then repeat the same process for calculating the independence score of Gram matrices from the content factors.
\begin{figure}[h]{}
    \centering
    \vspace{0mm}
    \includegraphics[width=0.25\textwidth]{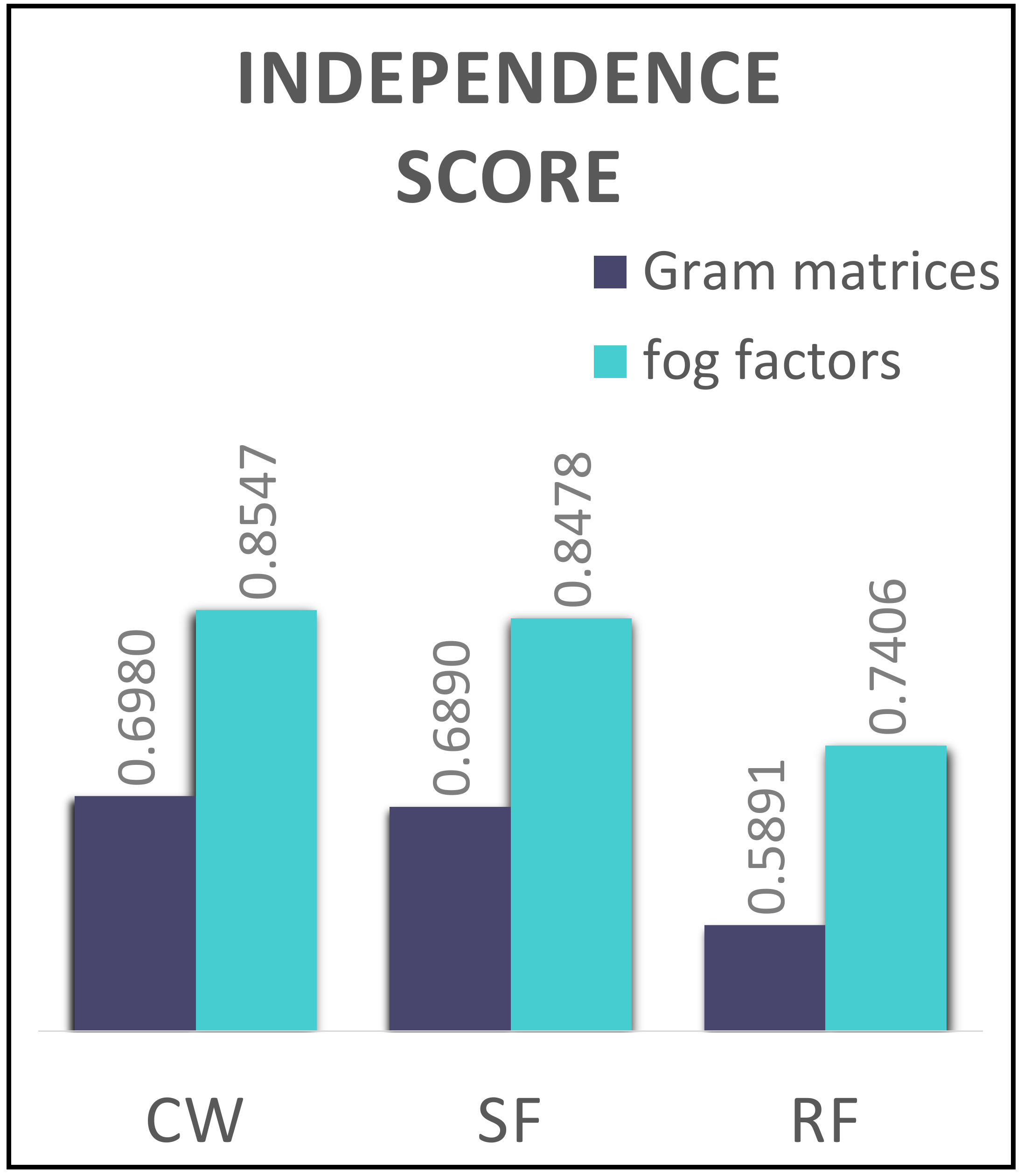}
\vspace{0mm}
\caption{Independence score of fog factors and Gram matrices from content factors. 
} 
\label{fig:content_factor_indepence}
\vspace{-1mm}
\end{figure}

Figure~\ref{fig:content_factor_indepence} presents the independent score of fog factors and Gram matrices from content factors.
Note that the experiment settings and dataset configurations are all the same as in the main paper.
Figure~\ref{fig:content_factor_indepence} proves that fog factors are more independent to content factors compared to Gram matrices, as desired.
It indicates that the fog-pass filtering module extracts only fog-relevant information apart from the image content.

\begin{table*}[ht]
\centering
\renewcommand{\arraystretch}{1.05}
\scalebox{1}{
\begin{tabular*}{2.1\columnwidth}{l|ccc|cccc}
\toprule
\multirow{2}{*}{Method}& \multicolumn{3}{c|}{~~Image Pair}                      & FZ & FDD       & FD     &C-Lindau   \\ 
                                & \sftype{CW} \& \sftype{SF}       & \sftype{CW} \& \sftype{RF}              &\sftype{SF} \& \sftype{RF}               & mIoU (\%)  & mIoU (\%) & mIoU (\%) & mIoU (\%)\\ \hline
\textbf{1 pair}               &                 &                      &              &            &           &   &        \\
Unidirectional (Fog to Clear) &                 & \sftype{CW} $\leftarrow$ \sftype{RF}  &              &  38.5   & 	36.3       &	45.6   &	67.1       \\ 
Unidirectional (Clear to Fog) &                 & \sftype{CW} $\rightarrow$ \sftype{RF}  &              &  36.6	 &  36.9	      & 46.2  & 	64.7        \\ 
Bidirectional                 &                 & \sftype{CW} $\leftrightarrow$ \sftype{RF}  &              & 37.7     & 40.3  & 47.2    & 66.0 \\
    \hline
\textbf{2 pairs }    &         &                 &              &              &            &           &           \\
Unidirectional (Fog to Clear)  &  \sftype{CW} $\leftarrow$ \sftype{SF}         &              & \sftype{SF} $\leftarrow$ \sftype{RF}              &   43.3	 & 39.2	  & 48.8     &	68     \\
Unidirectional (Clear to Fog)  &  \sftype{CW} $\rightarrow$ \sftype{SF}         &              & \sftype{SF} $\rightarrow$ \sftype{RF}  &   43.4	  &39.3	 &   48.4	  & 63.3    \\
Bidirectional                 & \sftype{CW} $\leftrightarrow$ \sftype{SF}         &              & \sftype{SF} $\leftrightarrow$ \sftype{RF}  & 46.0          & 47.6  & 50.0    & 62.3  \\
\hline
\textbf{3 pairs }   &           &                 &              &              &            &           &           \\
Unidirectional (Fog to Clear) & \sftype{CW}  $\leftarrow$ \sftype{SF}       & \sftype{CW} $\leftarrow$ \sftype{RF}       & \sftype{SF} $\leftarrow$ \sftype{RF}   & 44.4	 & 42.6	 & 47.1	  & 68.8  \\ 
Unidirectional (Clear to Fog)  & \sftype{CW} $\rightarrow$ \sftype{SF}       & \sftype{CW} $\rightarrow$ \sftype{RF}       & \sftype{SF} $\rightarrow$ \sftype{RF}   & 44.1	& 36.5	 & 46   &	64.5   \\   
Bidirectional (FIFO)                 & \sftype{CW} $\leftrightarrow$ \sftype{SF}       & \sftype{CW} $\leftrightarrow$ \sftype{RF}       & \sftype{SF} $\leftrightarrow$ \sftype{RF} & \textbf{48.4} & \textbf{48.9} & \textbf{50.7}  &  64.8 \\ 
\bottomrule
\end{tabular*}
}
\caption{
Analysis on the impact of the fog style matching loss.
\sftype{CW}, \sftype{SF}, and \sftype{RF} denote clear weather, synthetic fog, and real fog, respectively.
}
\label{tab:pair_directions}
\end{table*}

\section{Effect of Fog Style Matching Loss}
This section conducts extensive experiments to investigate the effect of the fog style matching loss $\mathcal{L}_{fsm}$.
In FIFO, the fog style matching loss $\mathcal{L}_{fsm}$ is carried out by bidirectionally matching each fog condition (\ie, \sftype{CW}, \sftype{SF}, and \sftype{RF}), so we denote it as a `Bidirectional' setting in this section.
We conduct additional experiments about variants of the fog style matching loss $\mathcal{L}_{fsm}$ in the `Unidirectional' setting (from Fog to Clear, from Clear to Fog).
For `Fog to Clear' settings, fog styles of real foggy images are unidirectionally matched to those of clear weather images, which is regarded as feature-level dehazing on real foggy images.
This is implemented simply by detaching the gradient flows from the fog style matching loss $\mathcal{L}_{fsm}$ to clear weather images.
For `Clear to Fog' settings, fog styles of clear weather images are unidirectionally matched to real foggy images similar to feature-level fog synthesis on clear weather images.
This is also implemented by detaching the gradient from the fog style matching loss $\mathcal{L}_{fsm}$ to real foggy images.

Table~\ref{tab:pair_directions} summarizes the results.
We found that the bidirectional fog style matching outperforms its unidirectional counterpart when the same domain pairs are involved; this result justifies the fog style matching loss in FIFO.
In addition, unidirectional (Fog to Clear) models have superior performance on the clear weather dataset~\cite{dai2020curriculum} compared to others due to the effect of focusing on clear weather conditions.

\section{Impact of Fog Factors}
\begin{figure}[ht]{}
    \centering
    \includegraphics[width=1\columnwidth]{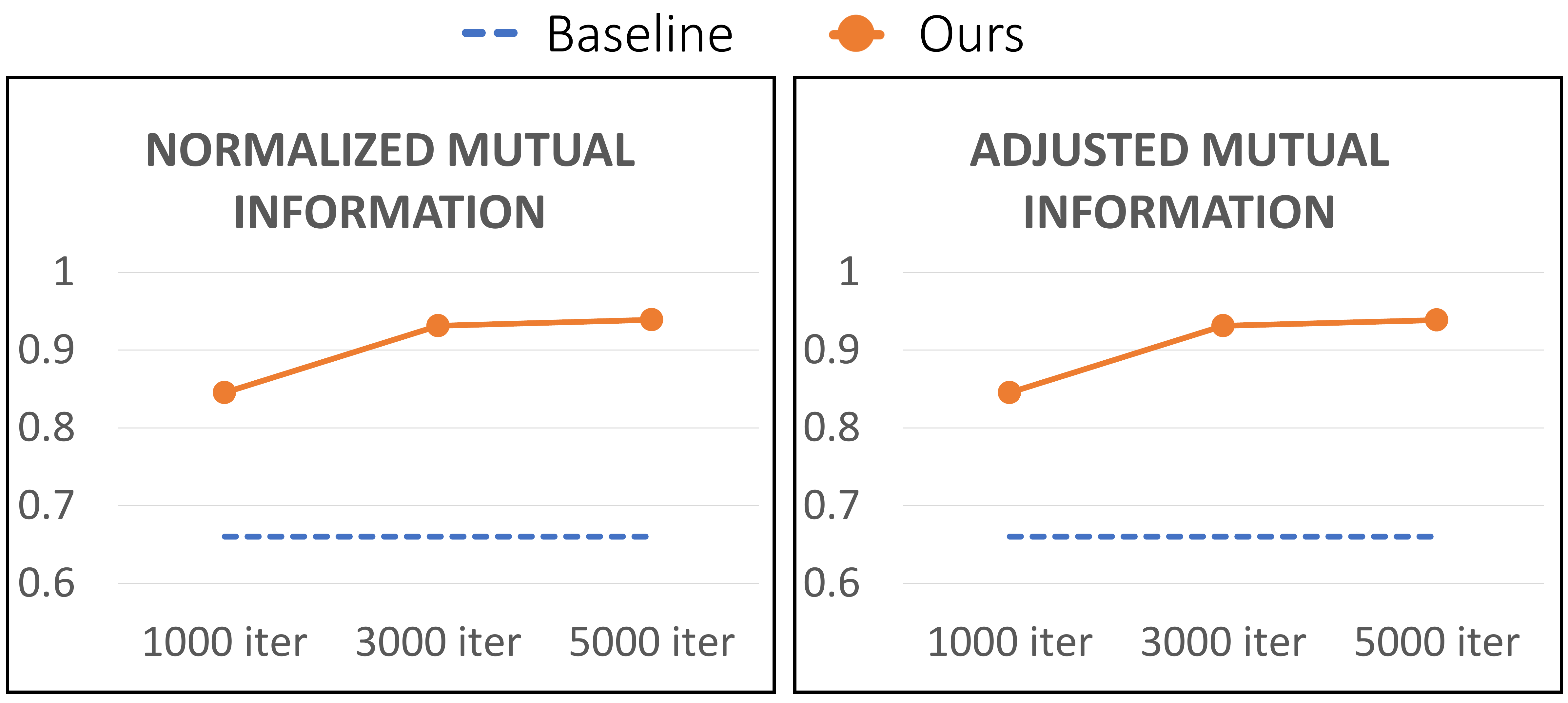}
\vspace{-6mm}
\caption{Comparison between fog factors and Gram matrices for the quality of $k$-means clustering by normalized mutual information and adjusted mutual information.
} 
\label{fig:kmeans}
\vspace{-1mm}

\end{figure}

We present additional comparison results for the quality of $k$-means clustering~\cite{hartigan1979algorithm} of the Gram matrices and that of the corresponding fog factors in other measures, normalized mutual information~\cite{estevez2009normalized}, and adjusted mutual information~\cite{vinh2010information}.
All of the measures prove the impact of fog factors in that they are more clustered than Gram matrices according to each fog condition as shown in Figure~\ref{fig:kmeans} and Table~\ref{tab:kmeans}.

\begin{table}[ht]
  \centering
    \centering
    \scalebox{1}{
\begin{tabular*}{0.8\columnwidth}{@{}C{2cm}@{}@{}C{1.5cm}@{}@{}C{1.5cm}@{}@{}C{1.5cm}@{}}
\toprule
 Comparison     & 1000 iter       & 	3000 iter	      & 5000 iter                         \\ \midrule
\multicolumn{4}{c}{\textit{Normalized Mutual Information}}   \\ \midrule
Gram matrix	           & 0.6602	             & 0.6602         & 0.6602          \\
Fog factor           & 0.8453	            & 0.9313       & 0.9387              \\ \midrule
\multicolumn{4}{c}{\textit{Adjusted Mutual Information}}   \\ \midrule
Gram matrix	           & 0.6601	              & 0.6601       & 0.6601        \\
Fog factor           & 0.8452          & 0.9313             & 	0.9387           \\
\midrule
\multicolumn{4}{c}{\textit{Adjusted Rand Index}}   \\ \midrule
Gram matrix	           & 0.6304	    & 0.6304	    & 0.6304	      \\
Fog factor           & 0.8683          & 0.9533    & 	0.9596            \\ 
\bottomrule
\end{tabular*}
}
    \vspace{-1mm}
    \caption{
Quantitative results of the quality of $k$-means clusters of the Gram matrices and fog factors in normalized mutual information, adjusted mutual information, and adjusted Rand index~\cite{hubert1985comparing}.
}
    \label{tab:kmeans}
    \vspace{0mm}

\end{table}

\section{Empirical Verification Using Evaluation Dataset}
\begin{figure}[ht]
\centering
\scalebox{0.95}{
\includegraphics[width=1\linewidth]{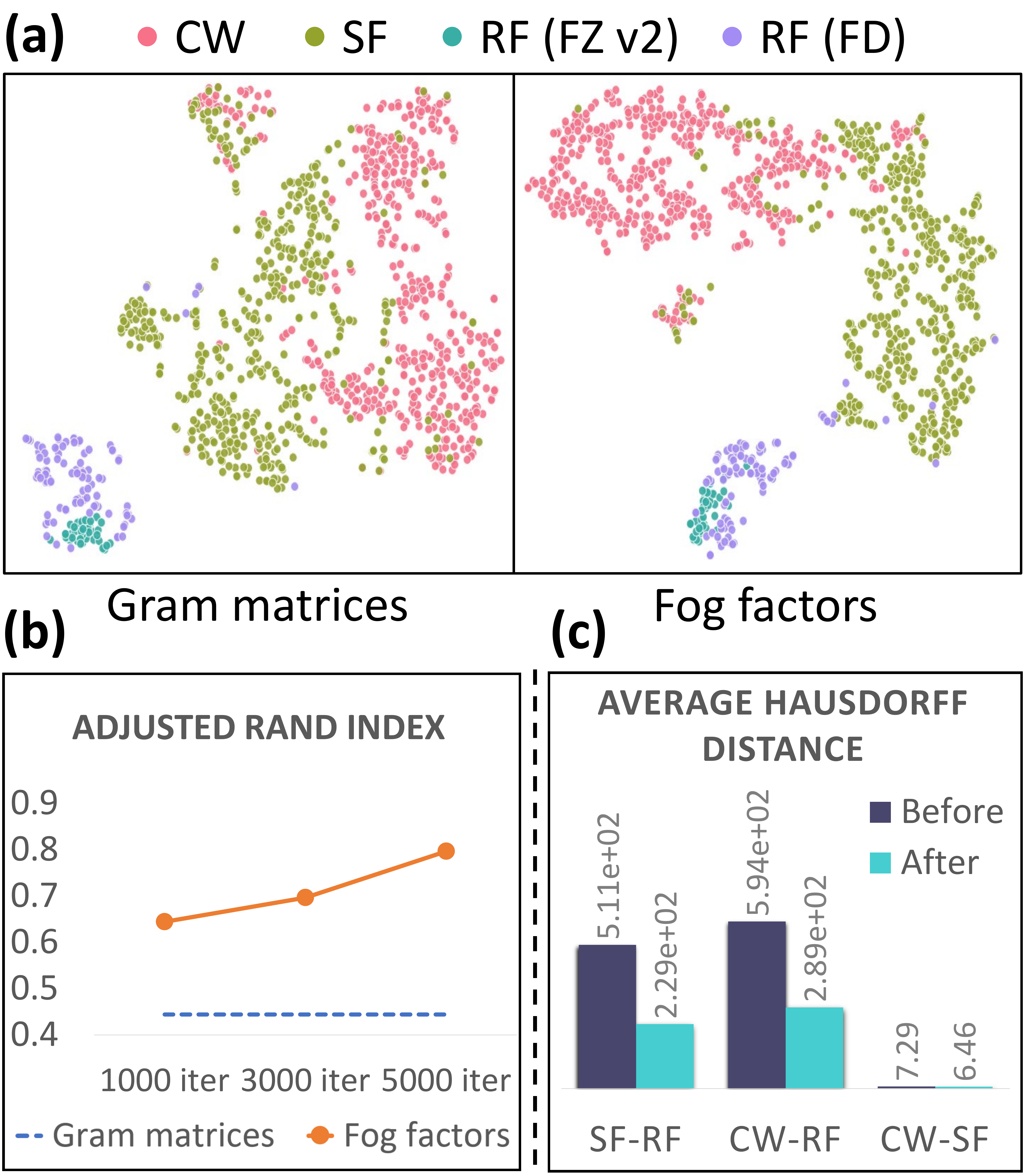}
}
\vspace{-1mm}
\caption{
Results of the empirical verification using the evaluation datasets.
(a) $t$-SNE visualization of distributions of Gram matrices and their fog factors.
(b) Comparison between the quality of $k$-means clustering of the fog factors and Gram matrices in adjusted Rand index.
(c) The fog-style gap between different domains before and after training with FIFO.}
\vspace{-0.8mm}
\label{fig:fz_empirical}
\end{figure}

We present additional results of the empirical verification using evaluation splits of the datasets, \ie, Cityscapes (500 images) as \sftype{CW}, Foggy Cityscapes-DBF (500 images) as \sftype{SF}, and Foggy Zurich-test v2 (40 images) and Foggy Driving (101 images) as \sftype{RF}.
\Fig{fz_empirical} shows that the tendency of the results is consistent with that of Fig.~4 of the main paper.

\section{Generalization on Deep Features}
\begin{figure}[ht]
\centering
\vspace{-1mm}
\scalebox{0.42}{
    \includegraphics[width=1.0\linewidth]{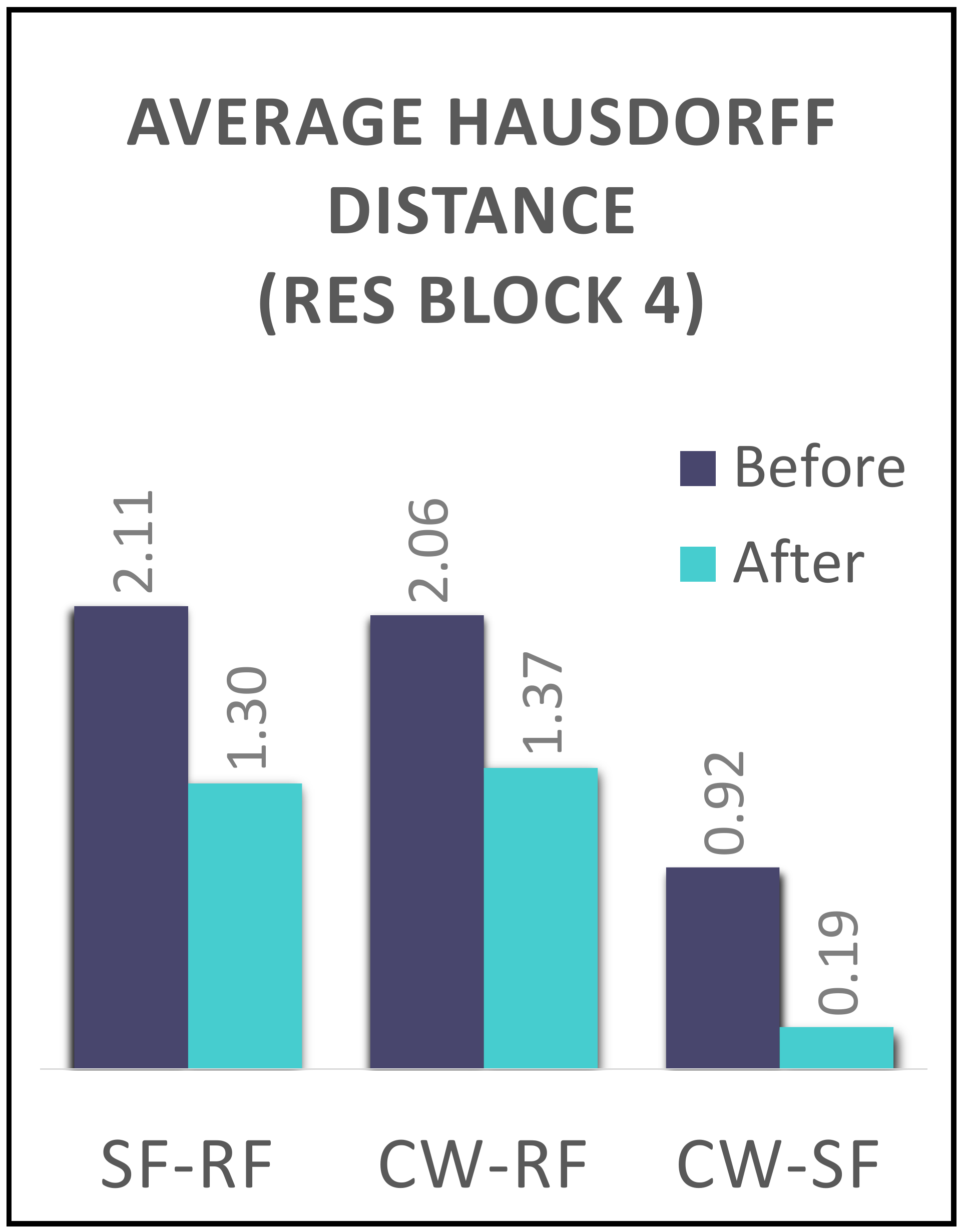}
}
\vspace{-1mm}
\caption{
Distances between sets of deep features from different domains before and after training with FIFO.}
\label{fig:deep_feature}
\end{figure}

This section empirically investigates the generalization of fog-invariant learning of our method on deep features.
It has been reported in domain adaptation and generalization literature~\cite{choi2021robustnet, zhou2021domain} that domain alignment at bottom layers closes the domain gap of deeper layer features.
As shown in \Fig{deep_feature}, we empirically verify our case: The average Hausdorff distances between ResBlock4 features from different domains decrease noticeably by FIFO. 

\section{Comparison with UDA}
In this section, we discuss the reason for failure when UDA methods are applied to the foggy scene segmentation in Table 1 of the main paper.

\begin{figure}[ht]
\centering
\vspace{0mm}
\scalebox{0.8}{
    \includegraphics[width=1\linewidth]{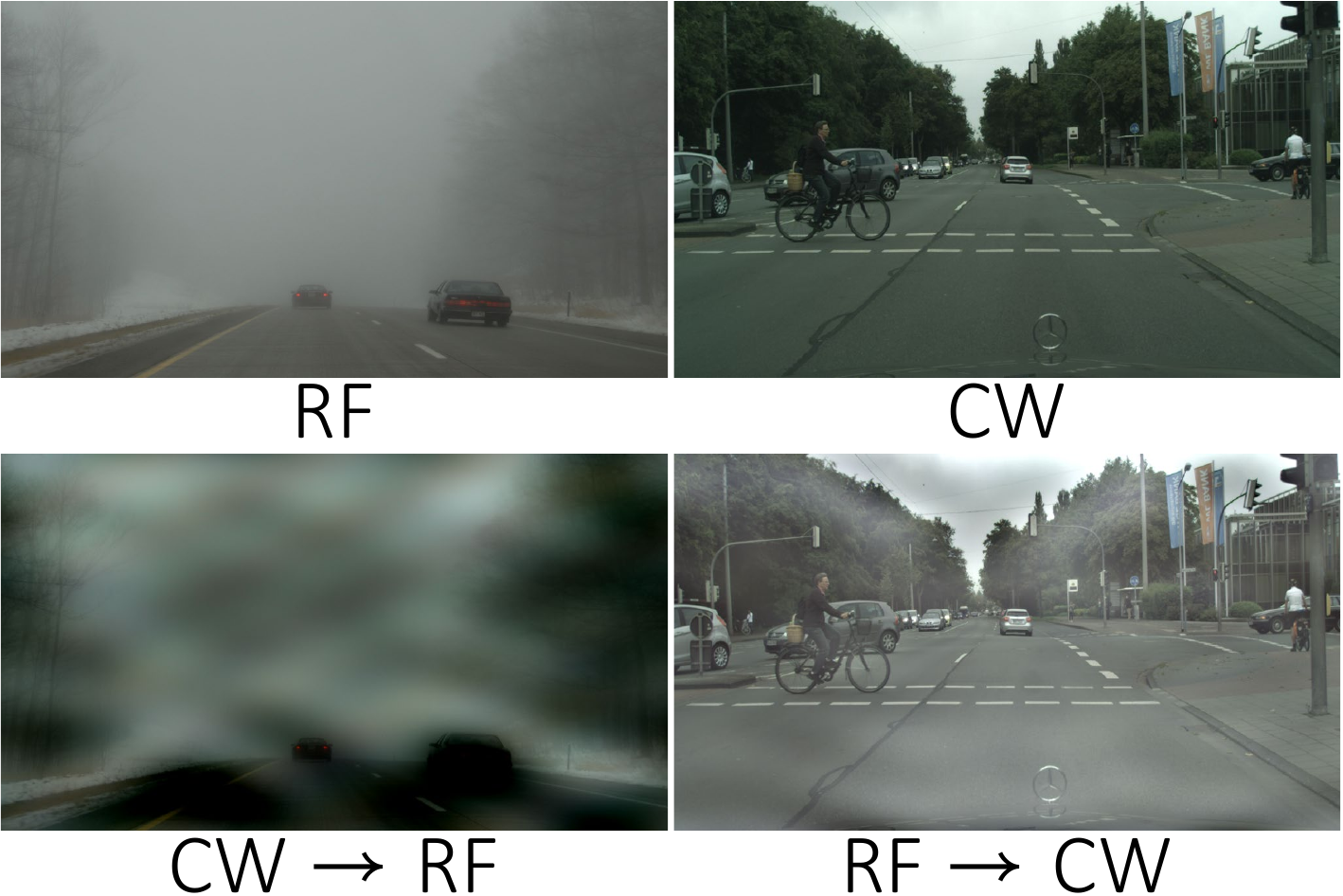}
}
\captionof{figure}{Outputs of spectral style transfer in FDA.}
\label{fig:fda}
\end{figure}

\begin{figure*}[t]
    \centering
    \includegraphics[width=1.0\textwidth]{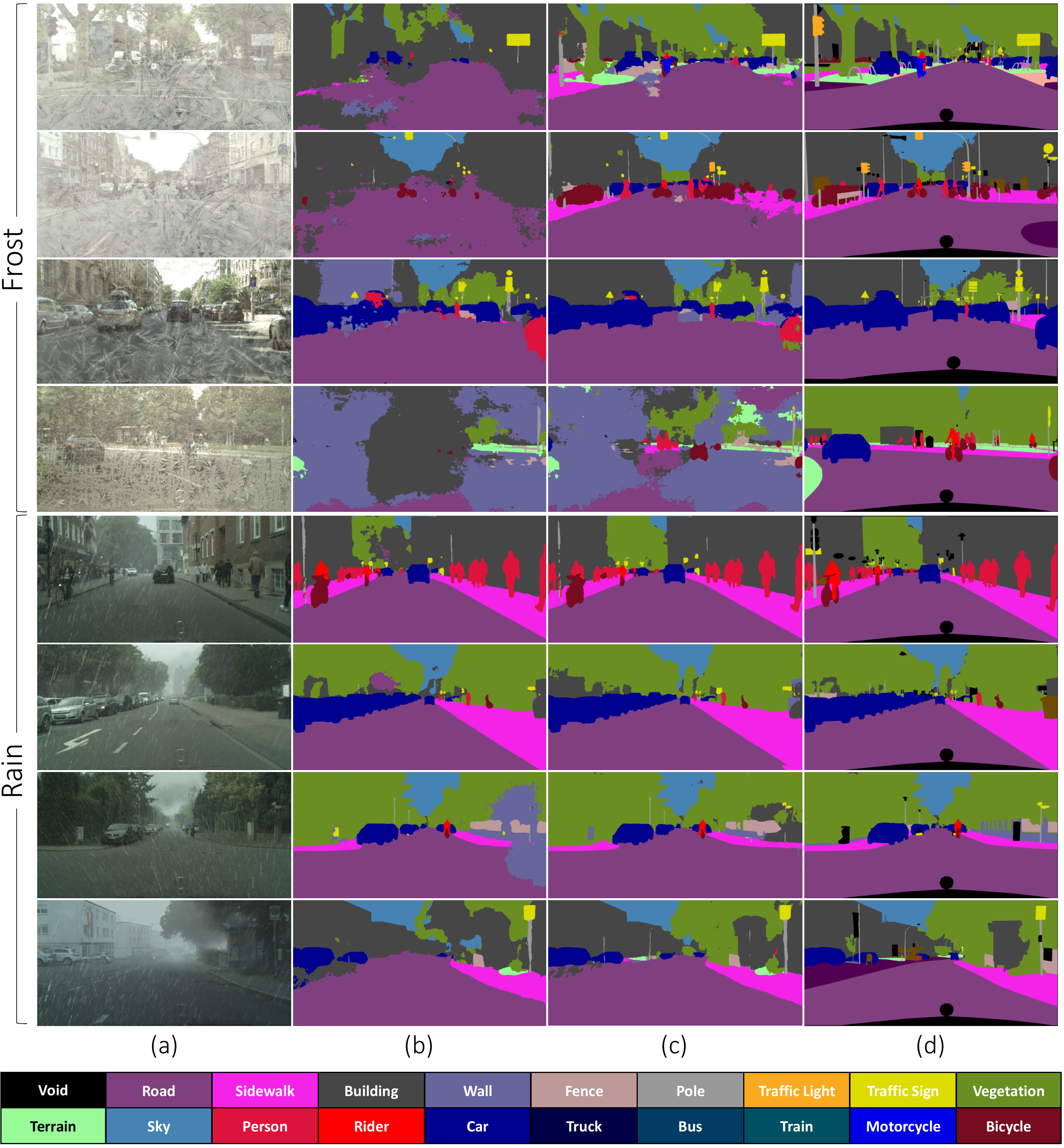}
    \vspace{-6.5mm}
\caption{
Additional qualitative results on \textit{frost} and \textit{rain} weather corruptions.
(a) Weather corrupted input images.
(b) Baseline.
(c) FIFO. (d) Groundtruth.
} 
\label{fig:weather_qual}
\vspace{0mm}
\end{figure*}

\noindent\textbf{Analysis on the failure of FDA.}
We suspect this is because the style representation of Fourier domain adaptation (FDA) is not suitable for handling foggy scenes:
FDA considers the low-frequency spectrum of an image as its style, but in the case of a foggy image, both its style and content lie in its low-frequency spectrum.
We verify this via qualitative results of the spectral style transfer, an intermediate step in FDA.
\Fig{fda} presents examples of the style transfer from \sftype{CW} to \sftype{RF} 
and from \sftype{RF} to \sftype{CW}.
The former causes severe artifacts on \sftype{RF} as the content \sftype{CW} as well as its style is transferred.
On the other hand, the latter applies fog effects to \sftype{CW}, but the result is not realistic.

\noindent\textbf{Superiority of FIFO over Domain Adversarial Learning.}
First, FIFO minimizes the entire objective function as presented in Section \ref{sec:algorithm} while domain adversarial learning optimizes the min-max loss. 
Hence, FIFO does not suffer from the instability issue of adversarial learning in training.
Second, FIFO can model and exploit within-domain fog style variations better than the domain adversarial learning (\eg, DANN). 
This is crucial since images of the same fog condition have different fog styles in general. 
FIFO achieves this property by the losses in Eq.~(1) and Eq.~(3) of the main paper;
the former motivated by metric learning enables the fog-pass filter to learn within-domain fog style variations, and the latter enables the segmentation model to keep such variations while closing style gaps only between different fog domains.
Accordingly, thanks to the superiority of FIFO over domain adversarial learning, FIFO clearly outperforms DANN in Table 1 of the main paper.

\section{Comparison with Variants of CMAda}
This section presents the comparison of FIFO to the variants of CMAda reported in~\cite{dai2020curriculum}, which suggests the current best performing model, CMAda3+.
Table~\ref{tab:compare_cmada} demonstrates the superiority of FIFO over the variants of CMAda.
In Table~\ref{tab:compare_cmada}, CMAda models denoted '+' are conducted the additional procedure of fog densification for making the fog density of real foggy training images similar to target fog density of the test real foggy images.
FIFO outperforms all variants of CMAda regardless of their number of stages and densification procedure.

\begin{table}[h]
  \centering
    \centering
    \renewcommand{\arraystretch}{1.05}
    \scalebox{1}{
\begin{tabular*}{0.85\columnwidth}{@{}C{2.5cm}@{}@{}C{1.5cm}@{}@{}C{1.5cm}@{}@{}C{1.5cm}@{}}
\toprule
method         & FZ test v2           & FDD             & FD                     \\ \midrule
CMAda1~\cite{dai2020curriculum}           & 38.9                & 36.6            & 46.0                   \\
CMAda2~\cite{Sakaridis_2018_ECCV}            & 42.9                 & 37.3            & 48.5          \\
CMAda3~\cite{dai2020curriculum}            & 43.7                 & 40.6            & 48.9          \\
CMAda2+~\cite{dai2020curriculum}            & 43.4                 & 40.1            & 49.9            \\
CMAda3+~\cite{dai2020curriculum}            & 46.8                 & 43.0            & 49.8            \\ \midrule
FIFO           & \textbf{48.4}                 & \textbf{48.9}            & \textbf{50.7}               \\\bottomrule
\end{tabular*}
}
    \vspace{-1mm}
    \caption{
Comparison of FIFO to variants of CMAda.
'+' denotes models applied the additional procedure of fog densification for real foggy training datasets.
The numbers attached to CMAda means the number of stages for curriculum learning.
}
    \label{tab:compare_cmada}
    \vspace{-3mm}

\end{table}

\section{Additional Qualitative Results} \label{sec:qualitative results}
\begin{figure*}[t]
    \centering
    \includegraphics[width=1.0\textwidth]{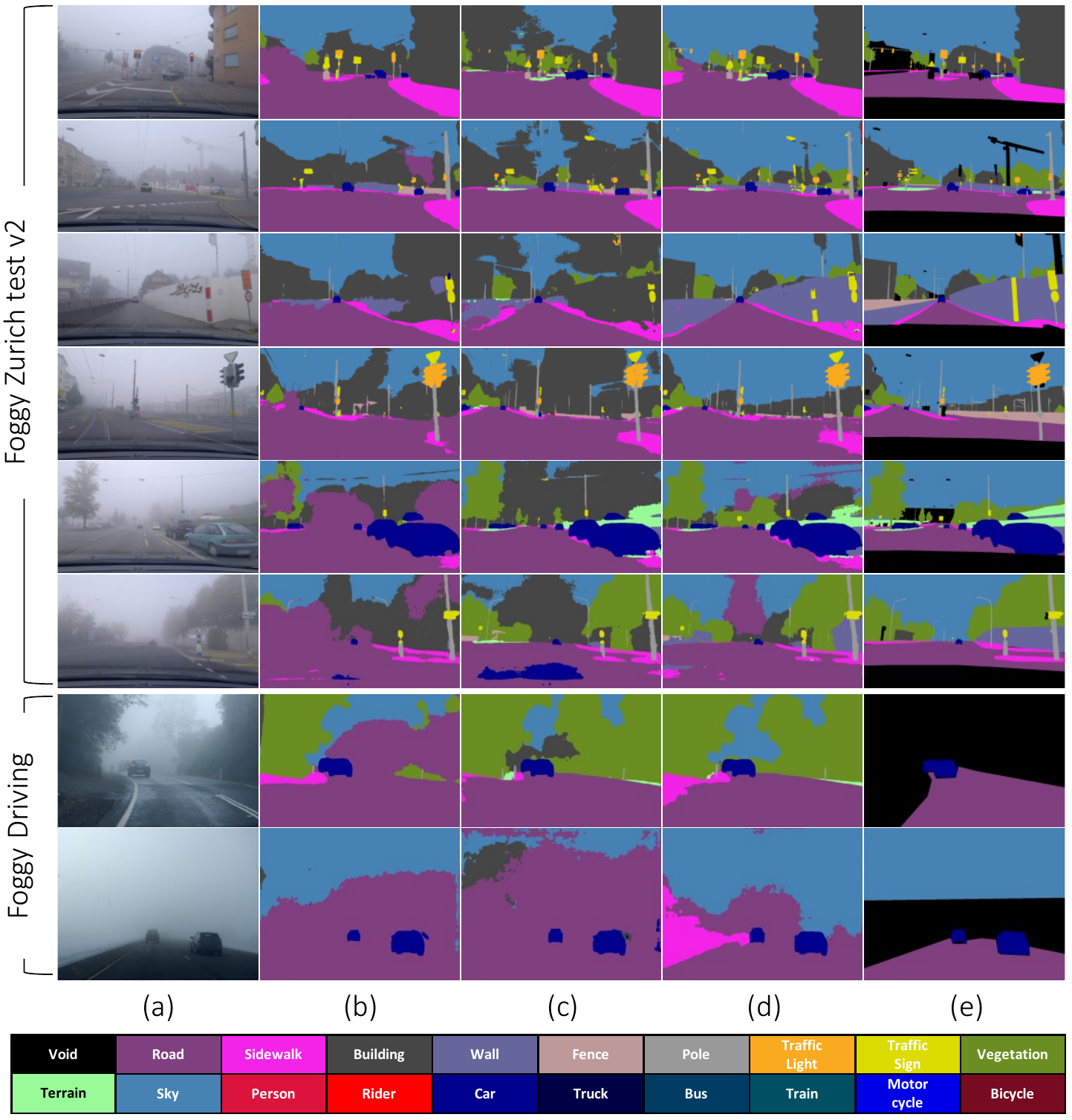}
\caption{
Additional qualitative results on the real foggy datasets.
(a) Real foggy images. 
(b) Baseline.
(c) Reduced version of FIFO closing the gap between gram matrices. (d) FIFO. (e) Groundtruth.
} 
\label{fig:more_qualitative_results}
\vspace{-3mm}
\end{figure*}

\begin{figure*}[t]
    \centering
    \includegraphics[width=1.0\textwidth]{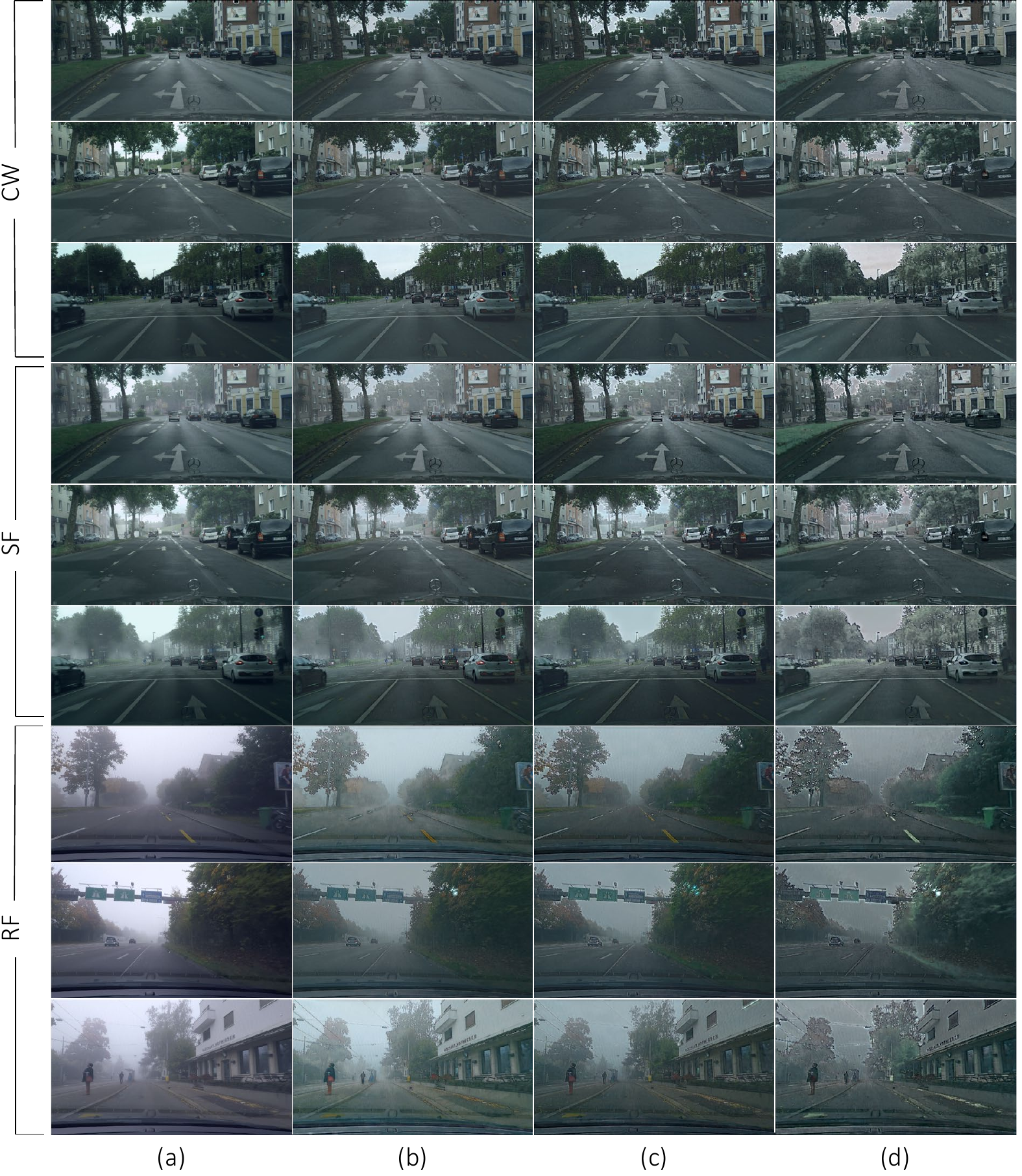}
\caption{
Additional qualitative results on image reconstruction.
(a) Real foggy images. 
(b) Baseline.
(c) Reduced version of FIFO closing the gap between gram matrices. (d) FIFO.
}  
\label{fig:more_recon_qualitative_results}
\vspace{-3mm}
\end{figure*}

This section presents additional qualitative results omitted in the main sections due to the space limit.
More segmentation results of FIFO are illustrated in Figure~\ref{fig:more_qualitative_results}.
We compare the results between FIFO, a variant of FIFO, by directly reducing the gap between Gram matrices and baseline.
Overall, FIFO offers higher quality segmentation results than the baseline regardless of fog density and datasets.
Specifically, FIFO seems best performing on parts where dense fog is laid while other models fail, which indicates FIFO working as desired.
Figure~\ref{fig:more_recon_qualitative_results} exihibits additional qualitative results on image reconstruction.
Likewise, the image quality where dense fog is laid is improved, which implies FIFO extract fog-invariant features.
In addition, clear weather images, as well as foggy images, become more clear when the features are trained by FIFO.

\clearpage

{\small
\bibliographystyle{ieee_fullname}
\bibliography{paper}
}

\end{document}